\documentclass[lettersize,journal]{IEEEtran}
\usepackage{amsmath,amsfonts}
\usepackage{amssymb} 
\usepackage{algorithmic}
\usepackage{algorithm}
 \usepackage{hyperref}
\usepackage{array}
\usepackage[caption=false,font=normalsize,labelfont=sf,textfont=sf]{subfig}
\usepackage{textcomp}
\usepackage{stfloats}
\usepackage{url}
\usepackage{verbatim}
\usepackage{graphicx}
\usepackage{cite}
\usepackage{xcolor}
\usepackage{enumitem}

\usepackage{booktabs,makecell,array}

\newcolumntype{L}[1]{>{\raggedright\arraybackslash}p{#1}}

\begin{document}

\title{A Survey on Cache Methods in Diffusion \\Models: Toward Efficient Multi-Modal Generation}

\author{
Jiacheng Liu$^{1}$,~
Xinyu Wang$^{1,2}$,~
Yuqi Lin$^{1}$,~
Zhikai Wang$^{1}$,~
Peiru Wang$^{1}$,~
Peiliang Cai$^{1}$,\\
Qinming Zhou$^{1,2}$,~
Zhengan Yan$^{1}$,~
Zexuan Yan$^{1}$,~
Zhengyi Shi$^{1}$,~
Chang Zou$^{1}$,~
Yue Ma$^{1,3}$,~
Linfeng Zhang$^{1\dag}$\\[4pt]
$^1$Shanghai Jiao Tong University ~~
$^2$Tsinghua University~~
$^3$The Hong Kong University of Science and Technology\\
\thanks{$^\dag$Corresponding author. Email: \texttt{zhanglinfeng@sjtu.edu.cn} }
\includegraphics[height=1.2em]{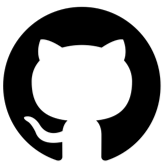}: \textbf{\href{https://github.com/Shenyi-Z/Cache4Diffusion.git}{\texttt{\textcolor{cyan}{Cache4Diffusion}}}} \\
\textbf{\href{https://github.com/Tammytcl/Awesome-Diffusion-Acceleration-Cache.git}{\texttt{\textcolor{cyan}{Awesome-Diffusion-Acceleration-Cache}}}}
}

\markboth{A Survey on Cache Methods in Diffusion Models: Toward Efficient Multi-Modal Generation}%
{Liu \MakeLowercase{\textit{et al.}}: A Survey on Cache Methods in Diffusion Models}


\maketitle

\begin{abstract}
Diffusion Models have emerged as a cornerstone of contemporary generative AI, owing to their exceptional generation quality and controllability. However, their inherent ``\textit{multi-step iterations}'' and ``\textit{complex backbone networks}'' inference paradigm results in prohibitive computational overhead and significant generation latency, which has become a critical bottleneck hindering their deployment in real-time interactive applications. Although existing acceleration techniques have made some progress, they often face challenges such as limited applicability, high training costs, or a degradation in generation quality.
Against this backdrop, Diffusion Caching presents a promising technical pathway as a training-free, architecture-agnostic, and efficient inference paradigm. Its core mechanism lies in accurately identifying and reusing the intrinsic computational redundancies within the diffusion inference process. Through feature-level cross-step reuse and inter-layer scheduling, it effectively reduces the computational load without altering the model parameters. This paper systematically reviews the theoretical foundations and technological evolution of Diffusion Caching and proposes a unified framework for its classification and analysis. 
Through a comparative analysis of representative methods, we indicate that Diffusion Caching exhibits a clear evolutionary trajectory from ``static reuse'' to ``dynamic prediction''. This trend not only enhances the flexibility of caching mechanisms in addressing the computational demands of diverse generation tasks but also holds great potential for deep integration with other mainstream acceleration techniques, such as sampling optimization and model distillation, to jointly construct a unified and efficient inference framework for future multimodal and interactive applications. Through this systematic review and forward-looking analysis, this paper aims to provide researchers with a clear technological roadmap for Diffusion Caching. We argue that this efficient inference paradigm will become a key enabling technology, driving the advancement of generative AI towards real-time performance and widespread adoption, thereby injecting new vitality into the theoretical construction and practical implementation of ``Efficient Generative Intelligence''. 
\end{abstract}



\section{Introduction}

\subsection{Background}

In recent years, Diffusion Models (DMs) have achieved groundbreaking progress in the field of generative artificial intelligence , particularly in image and video generation~\cite{cao2025hunyuanimage,wan2025}. They have emerged as the most representative generative paradigm following GANs~\cite{goodfellow2014generativeadversarialnetworks} and VAEs~\cite{kingma2022autoencodingvariationalbayes}. With the rise of the Diffusion Transformer (DiT)~\cite{peebles2023scalablediffusionmodelstransformers} architecture, new-generation models such as FLUX~\cite{blackforestlabs2024flux} and Qwen-Image~\cite{wu2025qwenimagetechnicalreport} in the field of image generation have shown strong generative capabilities. At the same time, in the field of video generation, a large amount of industrial investment has promoted the development of large-scale video diffusion models, producing several models with hundreds of billions of parameters, including open-source models such as Wan2.1~\cite{wan2025} and Hunyuan~\cite{kong2024hunyuanvideo}, and closed-source models such as Sora2.0~\cite{OpenAI2024VideoSimulators}, Movie-Gen~\cite{polyak2025moviegencastmedia}, and Seaweed~\cite{seawead2025seaweed}. These models have successively broken records in terms of generative quality, diversity, and controllability, showcasing remarkable generative potential.

\begin{figure*}[t]
  \centering
  \includegraphics[width=0.99\linewidth]{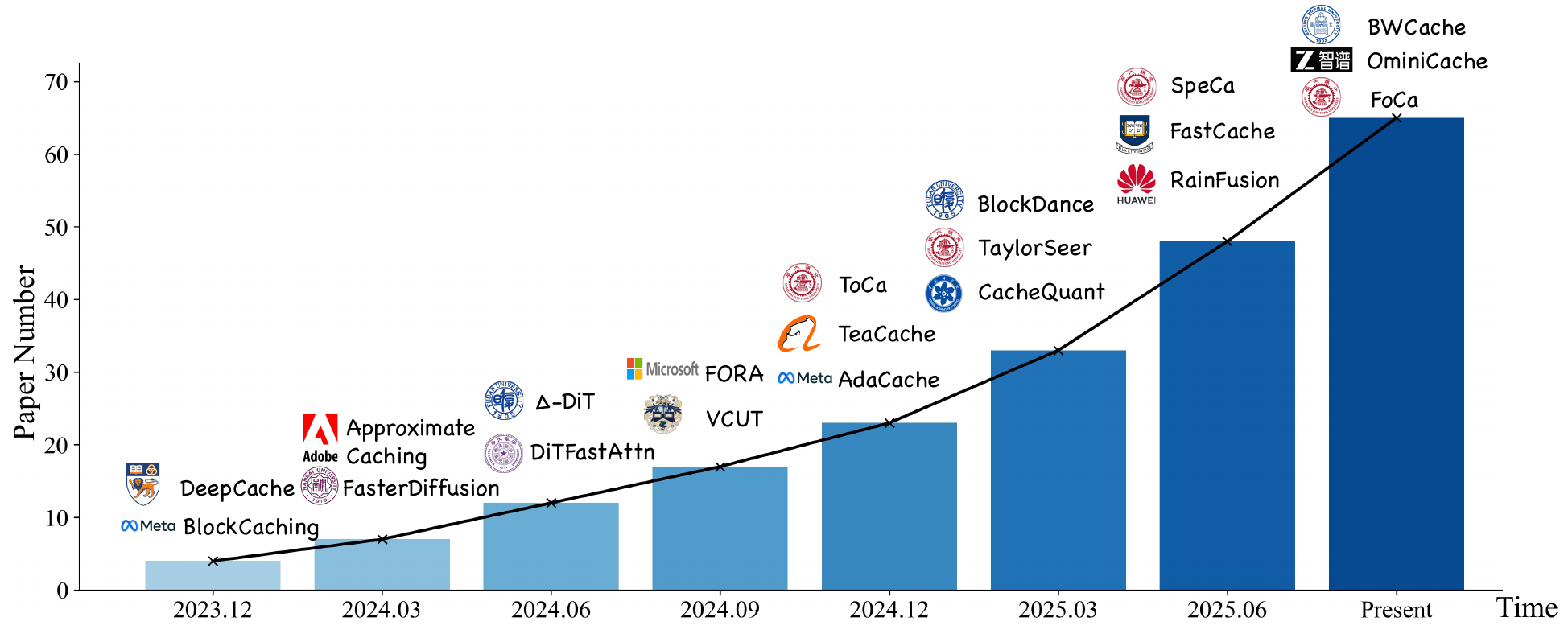}
  \caption{\textbf{The development trend of Diffusion Caching} The line chart illustrates the rapid growth in the number of related works from 2024 to the present. Representative works from each period are highlighted.}
  \label{fig:paper_count}
  \vspace{-10pt}
\end{figure*}

However, this leap in performance comes at the cost of rapidly escalating computational complexity and model scale. Owing to their ``s\textit{tep-by-step denoising}'' generation mechanism, diffusion models require multiple iterative forward passes through deep neural networks during sampling, making them inherently computation-intensive. This issue becomes particularly pronounced in high-resolution image generation and long-sequence video synthesis, where computational demands grow exponentially. For instance, generating a single $1328\times1328$ image with Qwen-Image~\cite{wu2025qwenimagetechnicalreport} involves approximately $1.29 \times 10^{4}$ TFLOPs of computation, resulting in a latency of up to 127 seconds per image on NVIDIA H20 GPU. Such substantial computational costs severely limit the feasibility of diffusion models in real-time creation and large-scale production applications.

Traditional acceleration efforts have primarily focused on optimizing numerical solvers~\cite{lu2022dpmsolver} and introducing model distillation techniques~\cite{salimans2022progressivedistillationfastsampling}. While these approaches can achieve partial speedup, they often struggle to balance acceleration and generation fidelity. Excessive reduction of sampling steps may lead to accumulated discretization errors, thereby degrading image quality. On the other hand, distillation and optimizing-based methods require additional computational and annotation resources, making them less flexible and harder to generalize across models.

Against this background, the high computational complexity and inference latency of diffusion models have become key bottlenecks for practical deployment. Generating a single high-resolution image can take tens of seconds to minutes, while producing a video may take several hours. This high latency not only reduces system throughput but also fails to meet the demands of real-time interaction and low-cost deployment. Therefore, achieving efficient inference without compromising generation quality has become a crucial challenge that must be addressed for diffusion models to transition from theoretical research to real-world applications.

\subsection{Challenges}

Despite the paradigmatic breakthroughs achieved by DMs in terms of generation quality and controllability~\cite{ma2025controllable}, their computationally intensive generative mechanism still faces severe efficiency bottlenecks during inference. The most prominent issue is the excessively long generation latency.

The inherent computational design of diffusion models, combining \textbf{\textit{multi-step iterations}} with \textbf{\textit{complex network architectures}}, is the primary source of latency.In generation process, the model need to progressively denoise high-dimensional noise, where each step involves extensive parameter computation and nonlinear mapping. High-fidelity synthesis typically requires tens of denoising steps (typically $20$ to $50$), and each step requires a full forward propagation through multi-layer deep networks such as U-Net or DiT. Therefore, inference time scales approximately linearly or even super-linearly with sequence length, image resolution, and network depth. For instance, on NVIDIA H20 GPU, generating a single 2K-resolution image often takes several minutes, while synthesizing a 720p video of around 129 frames may take several hours. As resolution and temporal length increase, generation latency grows exponentially. This problem is evident not only in academic benchmarks but also in practical applications.

Firstly, in interactive generation tasks, latency directly affects user experience. Real-time applications such as virtual dressing, real-time game scene generation, and live background replacement are extremely sensitive to response delays. When the latency between user input and model output reaches several seconds, or even minutes, the naturalness and immersion of interaction deteriorate dramatically, preventing the model from integrating effectively into real-time systems.

Second, when deploying diffusion models on on-device or edge systems, the limitation of computational resources further magnifies inference latency. Even with high-end consumer GPUs (e.g., NVIDIA RTX 4090), producing a high-resolution image can take tens of seconds, while mid- and low-tier hardware suffers from even more severe delays. Such strong reliance on hardware capabilities imposes a high entry threshold, hindering the scalability and accessibility of diffusion models in real-world applications.

Finally, in large-scale deployment on the cloud, generation latency directly affects the economic efficiency and scalability of services. The long inference time per generation request reduces system throughput and concurrency, forcing service providers to allocate additional computational resources to maintain responsiveness. This not only increases energy consumption and operational costs but also undermines the commercial viability of diffusion-based services.

Overall, high generation latency has become the primary bottleneck for diffusion models to move from laboratory research to practical applications. It not only limits the application of the model in real-time and interactive scenarios but also raises the economic threshold for large-scale deployment and hinders the popularization of the model on personal and edge computing devices. Therefore, how to efficiently reduce end-to-end generation latency—while maintaining generation quality and model stability, has become a central challenge in current research on accelerating diffusion model inference.

\subsection{Motivation}

To address the high inference latency of diffusion models, existing research primarily explores two technical directions: \textbf{(1) reducing the number of sampling steps }(\textit{Step Reduction}), and \textbf{(2) lowering the computational cost per-step} (\textit{Single-Step Cost Reduction}). Both directions theoretically aim to reduce the overall computational load during inference, yet in practice they must balance trade-offs among speed, generation quality, and adaptability, making it difficult to achieve an ideal equilibrium. This limitation provides a direct motivation for exploring new acceleration paradigms.

\begin{itemize}
    \item \textit{\textbf{Reducing the number of sampling steps}}

    This type of method usually models the diffusion process as a numerical integration problem of ordinary differential equations (ODE) or stochastic differential equations (SDE), and reduces the number of sampling times by expanding the integration step length through \textbf{high-order numerical solvers}, such as DPM-Solver~\cite{lu2022dpmsolver}, UniPC~\cite{zhao2023unipcunifiedpredictorcorrectorframework}. The core idea is to simulate the denoising trajectory with larger time steps to achieve sampling compression. However, when the number of sampling steps is reduced to a certain threshold (for example, less than 10 steps), discretization errors accumulate rapidly, leading to missing details, structural distortion, and visual artifacts in the generated results. In addition, another type of \textbf{distillation method}, such as consistency distillation, progressive distillation, approximates the multi-step behavior of the original model by training lightweight models, thus completing generation in fewer steps. Although these methods can achieve higher acceleration, they have high training costs, limited adaptability, and are sensitive to the target model and task conditions, which are not conducive to general deployment.
    \item \textit{\textbf{Lowering the computational cost per-step} }

    This direction focuses on reducing the computational burden of each forward pass through model compression or system-level optimization. Model compression techniques such as \textbf{quantization}~\cite{jacob2017quantizationtrainingneuralnetworks} and \textbf{pruning}~\cite{han2015learningweightsconnectionsefficient} can lower computation and memory costs but often sacrifice generation quality and require complex retraining procedures. \textbf{Lightweight architecture design} can reduce computation structurally, but it is constrained by limited model capacity and dependence on pretrained ecosystems, making it difficult to maintain both fidelity and generalization. Meanwhile, \textbf{system-level optimization techniques}, such as FlashAttention~\cite{dao2022flashattentionfastmemoryefficientexact}, TensorRT acceleration, improve operator efficiency and memory scheduling to enhance hardware utilization. While they can yield performance gains, these methods rely heavily on specific platforms and hardware support, which limits their portability and generality. Moreover, their performance is often sensitive to input distribution and model architecture, increasing the complexity of system tuning and maintenance.
\end{itemize}


In this context, \textbf{\textit{Diffusion Caching}} emerges as a promising optimization paradigm based on a different principle. Unlike methods that modify model architecture or require retraining, caching aims to identify and eliminate computational redundancy during inference. In the iterative denoising process of diffusion models, many intermediate results can be reused. For example, under fixed text conditions, the key and value matrices in Cross-Attention layers remain constant across timesteps, while feature maps between adjacent timesteps often change slowly. Recomputing these redundant results leads to substantial waste. \textit{\textbf{Diffusion Caching} leverages such temporal correlation and invariance to enable computational reuse and efficient inference}.

This concept has been well validated in autoregressive generation with large language models (LLMs), where KV-Cache has become a standard component, providing multi-fold acceleration without loss of quality. By analogy, caching in diffusion models offers two core advantages:  
 \textit{\textbf{(1)Training-free nature}}: \textit{Diffusion Caching} operates purely at inference time, requiring no additional training or fine-tuning;  
\textit{\textbf{(2) Orthogonality and composability}}: it can be combined with other acceleration techniques, such as step reduction or model compression, to achieve complementary benefits.

Therefore, \textit{\textbf{Diffusion Caching}} should not be viewed as a replacement for existing acceleration strategies but rather as a low-cost, highly compatible, and easily deployable optimization technique. It offers a new research direction and theoretical foundation for overcoming the inference bottlenecks of diffusion models without compromising generation quality.

\subsection{Contributions}

Currently, this field lacks a comprehensive survey and unified theoretical framework, and significant differences remain across existing studies in terms of principle interpretation, strategy design, and applicability. To fill this gap, the main contributions of this work are as follows:

\begin{enumerate}[leftmargin=1.2em, labelsep=0.3em]
    \item \textbf{\textit{The first systematic summary of the theory and practice of diffusion caching.}} 
    We explain the core idea of diffusion caching from a principled perspective: \textit{identifying and reusing computational redundancies in the diffusion inference process to reduce repeated computation and achieve acceleration}. We interpret the mathematical essence, applicable conditions, and constraints of caching from the dual perspectives of numerical analysis and neural network computation graphs.  By reviewing existing research, we clarify the unique position of diffusion caching within the diffusion model acceleration ecosystem, providing a unified theoretical framework for future studies.

    \item \textbf{\textit{A unified taxonomy and analytical framework for diffusion caching.}} 
    We construct the first systematic classification and analytical framework for diffusion caching along three dimensions: trigger condition, reuse granularity, and update strategy, revealing the intrinsic logic and technological evolution among different methods. We categorize existing approaches into two levels: \textbf{Static} and \textbf{Dynamic Caching}. Within the dynamic caching paradigm, we further identify four representative strategies: \textbf{Timestep-Adaptive}, \textbf{Layer-Adaptive}, \textbf{Predictive}, and \textbf{Hybrid Caching}. Through systematic comparison and analysis, we reveal the technological trajectory of diffusion caching from \textit{``Static Reuse''} to \textit{``Dynamic Prediction.''}. 

    \item \textbf{\textit{Evolution trends and future research directions of diffusion caching.}} 
    Building upon a comprehensive review of existing work, we summarize common challenges and potential breakthroughs in diffusion caching. Existing methods still face limitations in terms of \textbf{cache consistency}, \textbf{generalization}, \textbf{memory efficiency}, and c\textbf{ross-platform adaptability}, particularly in high-resolution image and long-sequence video generation tasks. Looking forward, we anticipate multi-dimensional development trends: on one hand, \textit{diffusion caching} mechanisms can be combined with other inference techniques to achieve multi-level acceleration; on the other hand, the concept of \textit{diffusion caching} can be extended to complex generation tasks, providing new scalable avenues for efficient generation.
\end{enumerate}

In summary, this paper not only provides \textbf{\textit{the first systematic overview}} of diffusion caching from theoretical, methodological, and application perspectives but also proposes a unified analytical framework and future directions. We believe that diffusion caching will become a cornerstone technology for future Efficient Generation, offering a new research paradigm and engineering perspective for practical deployment and sustainable optimization of diffusion models.

\section{Related Methods}

\subsection{Distillation}

\textbf{\textit{Distillation methods }}accelerate diffusion sampling by \textit{training a student model to approximate the teacher’s denoising process, compressing the long iterative trajectory into only a few or even a single step}. Progressive Distillation~\cite{salimans2022progressive} reduces hundreds of sampling steps to several through iterative halving with minimal quality loss. Consistency Models~\cite{song2023consistency} enforce prediction agreement across timesteps, enabling one-step or few-step high-quality sampling. GD-Distill~\cite{meng2023guided} integrates classifier-free and non-classifier-free diffusion into the student model to reduce computation. LCM~\cite{luo2023latentconsistencymodelssynthesizing} performs distillation in latent space, achieving high-quality results within 1–4 steps. MCM~\cite{heek2024multistep} unifies diffusion and consistency frameworks, supporting flexible-step sampling with improved stability. RG-LCM~\cite{wu2024rglcm} incorporates human preference signals during distillation, surpassing the teacher in perceptual quality for one-step generation.

In video generation, ADL~\cite{yang2024animatediff} fuses multiple teacher models to accelerate synthesis. MoCM~\cite{zhao2024mocm} and AVDM²~\cite{xu2024dmdistill} enhance temporal and visual coherence through motion–appearance disentanglement and distribution-matching objectives. VIP~\cite{sun2025vip} and DAP~\cite{huang2025dap} introduce preference-guided or adversarial optimization, enabling one-step video generation with improved fidelity and diversity.


\subsection{Pruning \& Sparsification}

\textbf{\textit{Pruning}} and \textbf{\textit{sparsification}} reduce computational and memory cost by \textit{removing redundant parameters, sparsifying weights, or compressing intermediate representations}. DC~\cite{han2016deepcompression} performs joint compression via pruning, quantization, and coding, greatly lowering parameter and bandwidth requirements with minimal quality loss. PCNN~\cite{li2017pruningcnn} accelerates inference through channel- and filter-level pruning, while LTH~\cite{frankle2019lottery} reveals trainable sparse subnetworks within over-parameterized models, inspiring structured and unstructured compression in later diffusion architectures. RTL~\cite{chen2020rigging} improves sparse network trainability through reset-and-retrain strategies, providing a foundation for modern sparse diffusion designs.

In image generation, research primarily focuses on compressing U-Net and DiT architectures. SPD~\cite{fang2023spd} applies structured pruning to reduce channels and attention heads. TM-SD~\cite{hu2023tmsd} and TFusion~\cite{li2023tfusion} merge or fuse tokens to reduce self-attention cost, while DFA~\cite{li2024dfa} targets attention computation optimization in DiT-based models. IBTM~\cite{liu2024ibtm} and ToMA~\cite{wang2025toma} employ token-merging mechanisms to enhance efficiency for both image and video generation. DC-Gen~\cite{he2025dcgen} connects a deeply compressed autoencoder with diffusion models, reducing the redundancy in latent space.

Studies for video emphasize reducing spatiotemporal complexity. VSA~\cite{zhang2025vsa} introduces trainable sparse attention to lower inter-frame computation, and SvDiT~\cite{liu2025svdit} incorporates sparse spatiotemporal attention for efficient video synthesis. MegaTTS-3~\cite{li2025megatts3} applies sparse alignment to strengthen latent DiT representations for zero-shot speech generation. DC-VideoGen~\cite{chen2025dc} extends DC-Gen in video generation, reducing the number of latent space tokens while preserving high reconstruction quality.


\subsection{Parallel Sampling \& Hardware Acceleration}

\textbf{\textit{Parallel acceleration methods}} improve efficiency by \textit{introducing concurrency at the operator, timestep, or task level, minimizing redundant memory access and transforming the inherently sequential sampling process into parallel execution.} This greatly reduces latency and increases throughput. FA~\cite{dao2022flashattention} minimizes memory I/O overhead through I/O-aware on-chip tiling, while FA2~\cite{dao2024flashattention2} further optimizes workload partitioning and scheduling, approaching the theoretical efficiency limit of matrix multiplication.

In parallel sampling, SpecDiff~\cite{christopher2025specdiff} performs draft–verify generation by producing multiple candidates simultaneously, avoiding strictly sequential sampling. SpecSampling~\cite{debortoli2025specsampling} extends this idea to continuous diffusion, enabling fully parallelized inference without auxiliary draft models. APD~\cite{israel2025apd} dynamically allocates parallel computation during decoding, effectively lowering latency and improving efficiency. STADI~\cite{liang2025stadi} introduces step–block partitioning and scheduling in heterogeneous GPU environments to achieve balanced workload distribution.

At the hardware and system level, DF~\cite{li2024distrifusion} achieves near-linear multi-GPU scaling for high-resolution generation through image partitioning and asynchronous communication. PF~\cite{fang2024pipefusion} combines inter-layer pipelining with block-level parallelism to enable efficient inference in Diffusion Transformers. RAIN~\cite{shu2024rain} introduces frame-level parallel attention and cross-temporal noise-layer updates, efficiently supporting continuous and long-duration video generation.


\subsection{Sampler Optimization}

\textbf{\textit{Sampler optimization methods}} aim to \textit{reduce the number of iterations during the reverse inference phase of diffusion models through improved numerical integration and timestep scheduling}. DDPM~\cite{ho2020ddpm} establishes the foundational denoising framework using variational inference to model the reverse Markov chain. DDIM~\cite{song2021ddim} introduces deterministic sampling by reformulating the process into a non-Markovian form, enabling faster high-quality generation. SDE~\cite{song2021sde} unifies diffusion under stochastic differential equations, and IDDPM~\cite{nichol2021iddpm} further incorporates learnable variance and refined timestep scheduling to enhance stability in few-step sampling.

Building on these foundations, PNDM~\cite{liu2022pndm} applies pseudo-numerical integration to maintain stability with limited steps, while DPM-Solver~\cite{lu2022dpmsolver} and DPM-Solver++~\cite{lu2022dpmsolverpp} employ high-order ODE solvers to achieve high-quality results within about ten iterations. UniPC~\cite{zhao2023unipc} unifies prediction–correction and multi-step solvers into a general framework for accurate few-step generation. Recent works such as RAPID³~\cite{chen2025rapid3} and S4S~\cite{kim2025s4s} introduce learning-based and reinforcement-driven optimization, jointly tuning step size, solver order, and scheduling to enable end-to-end adaptive sampling with minimal steps.


\subsection{Cache Acceleration}

Although methods such as distillation, pruning and sparsification, parallel acceleration, and sampler optimization have substantially reduced the inference cost of diffusion models, each still faces intrinsic limitations. \textbf{\textit{Distillation}} requires extensive retraining on teacher-generated trajectories, demanding high computational cost and often leading to quality degradation under extremely few-step sampling due to insufficient distribution coverage. \textbf{\textit{Pruning}} and \textbf{\textit{sparsification}} mainly reduce per-step computation, but the iterative denoising process remains unchanged, leaving overall latency largely unimproved. Moreover, structured pruning and token sparsification can weaken representational capacity, causing semantic misalignment or loss of fine details under high compression ratios. \textbf{\textit{Parallel and hardware-based acceleration methods}} depend heavily on specific architectures, kernel implementations, and communication strategies. While they achieve near-linear scaling on multi-GPU systems, their benefits diminish under limited hardware, and large-scale parallelism often introduces synchronization overhead and consistency issues across timesteps or regions. \textbf{\textit{Sampler optimization}} reduces the number of steps through advanced numerical solvers and scheduling, yet still fundamentally relies on iterative integration, making computational cost scale linearly with sequence length or resolution. These approaches also tend to be sensitive to guidance strength, error accumulation, and distribution shifts, limiting their robustness and general applicability.

To overcome these constraints, \textbf{\textit{Diffusion Caching}} introduces a complementary perspective by \textit{explicitly storing and reusing intermediate activations or tokens during inference}. Without additional training or architectural modification, it directly avoids redundant computations across timesteps, thereby mitigating the inherent inefficiency of stepwise iteration.

\begin{table*}[t]
\centering
\caption{Comparison of Diffusion Model Acceleration Techniques.}
\renewcommand{\arraystretch}{1.15}

\resizebox{\textwidth}{!}{
\begin{tabular}{
>{\centering\arraybackslash}m{0.1\textwidth} 
>{\centering\arraybackslash}m{0.24\textwidth} 
>{\centering\arraybackslash}m{0.26\textwidth} 
>{\centering\arraybackslash}m{0.2\textwidth} 
>{\centering\arraybackslash}m{0.26\textwidth}}
\toprule
\textbf{Category} & 
\makecell{\textbf{Core} \textbf{Mechanism /}\\\textbf{Technique}} &
\makecell{\textbf{Representative}\\\textbf{Work}} &
\textbf{Advantages} & 
\textbf{Limitations} \\
\midrule

\textbf{Distillation} & 
Student model learns teacher’s denoising mapping to compress long sampling trajectories. &
GD-Distill~\cite{meng2023guided}, 
LCM~\cite{luo2023latentconsistencymodelssynthesizing}, 
MCM~\cite{heek2024multistep}, 
RG-LCM~\cite{wu2024rglcm}, 
DAP~\cite{huang2025dap}. &
Reduces sampling steps dramatically while maintaining quality; flexible for image/video generation. &
Requires costly retraining; quality drops at extreme few-step generation; weak cross-domain generalization. \\

\midrule
\textbf{Pruning \& Sparsification} & 
Remove redundant parameters or weights; structured/unstructured sparsity in U-Net or DiT. &
SPD~\cite{fang2023spd}, 
TM-SD~\cite{hu2023tmsd, li2023tfusion}, 
VSA~\cite{zhang2025vsa}, 
DFA~\cite{li2024dfa}, 
ToMA~\cite{wang2025toma}. &
Reduce redundant parameters and computation with minimal quality loss; compatible with existing architectures. &
Limited real speedup due to hardware sparsity support; potential loss of fine details or temporal stability; task-specific robustness; incompatible with efficient attention variants; cannot mitigate iterative sampling latency. \\

\midrule
\textbf{Parallel Acceleration} & 
Introduce concurrency at operator/timestep/task level; multi-GPU and pipelined execution. &
FA2~\cite{dao2024flashattention2}, 
SpecDiff~\cite{christopher2025specdiff}, 
SpecSampling~\cite{debortoli2025specsampling}, 
APD~\cite{israel2025apd}, 
STADI~\cite{liang2025stadi}. &
Significantly reduce inference latency through kernel optimization, distributed scheduling, and speculative decoding; achieve real-time generation on high-end GPUs. &
Strong hardware dependency and poor portability; communication overhead scales poorly with multi-GPU setups; limited benefit on single-GPU systems; parallelization may break temporal/spatial consistency and cause quality instability. \\

\midrule
\textbf{Sampler Optimization} & 
Optimize numerical integration and timestep scheduling to reduce diffusion iterations. &
PNDM~\cite{liu2022pndm}, 
DPM-Solver~\cite{lu2022dpmsolver}, 
UniPC~\cite{zhao2023unipc}, 
RAPID³~\cite{chen2025rapid3}, 
S4S~\cite{kim2025s4s}. &
Greatly reduce sampling steps while preserving quality; broadly applicable to existing diffusion models. &
Limited by iterative nature; sensitive to score and guidance variations; may lose details or stability under few steps; weak cross-model robustness. \\

\bottomrule
\end{tabular}}
\end{table*}




\section{Taxonomy of Cache Acceleration}
\begin{figure*}[t]
  \centering
  \includegraphics[width=0.99\linewidth]{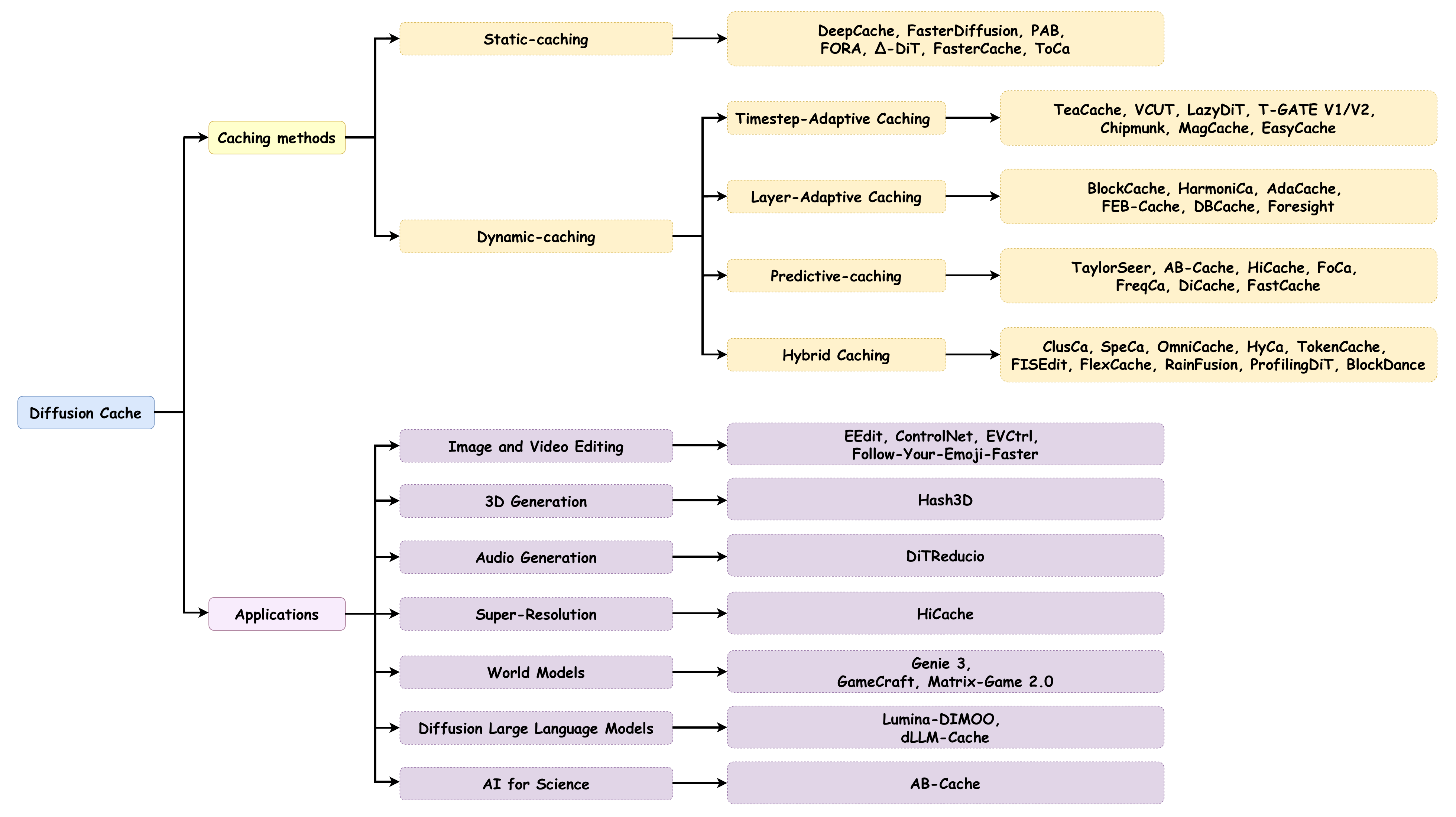}
  \caption{\textbf{Overview of Diffusion Cache Framework} The upper panel illustrates our unified taxonomy that categorizes caching methods into Static and Dynamic paradigms, with Dynamic Caching further divided into Timestep-Adaptive, Layer-Adaptive, Predictive, and Hybrid strategies. The lower panel shows applications across diverse generative tasks.}
  \label{fig:framework}
  \vspace{-10pt}
\end{figure*}
\subsection{Diffusion Theory}

Diffusion models have emerged as one of the most influential generative models in recent years, with their core concepts rooted in the physical diffusion process. These models simulate the random diffusion phenomenon of particles in thermodynamics, constructing a forward diffusion and reverse generation paradigm that achieves exceptional performance in multimodal content generation tasks including images \cite{rombach2022highresolutionimagesynthesislatent}, videos, and audio. To understand the mechanisms of subsequent caching acceleration techniques, this section elucidates the theoretical foundations of diffusion models.

The core idea of diffusion models can be summarized as \textit{a bidirectional stochastic process consisting of the forward diffusion process and the reverse generation process}. In \textbf{the forward diffusion process}, given a real data distribution $q(x_0)$, data samples $x_0$ are transformed into pure noise $x_T$ through the gradual addition of Gaussian noise. This process can be understood as gradually converting complex data distributions into simple standard Gaussian distributions until the structural information of the original data is completely masked by noise. \textbf{The reverse generation process} is the temporal inversion of the forward process, which learns a neural network to progressively recover meaningful data structure from pure noise. This process learns how to remove noise, achieving a mapping from simple distributions to complex data distributions. This design offers key advantages. The forward process provides a clear training objective through noise prediction. Meanwhile, the reverse process uses learnable neural networks to approximate the denoising distribution, avoiding the complexity of adversarial training \cite{goodfellow2014generativeadversarialnetworks} or variational inference \cite{kingma2022autoencodingvariationalbayes} in traditional generative models.

In mathematical formalization, the forward diffusion process is modeled as a fixed Markov chain, where each step adds a small amount of Gaussian noise to the data. Given a data point $x_0 \sim q(x_0)$, the forward process is defined as

\begin{equation}
q(x_{1:T}|x_0) := \prod_{t=1}^{T} q(x_t|x_{t-1}),
\end{equation}
where the transition kernel is
\begin{equation}
q(x_t|x_{t-1}) = \mathcal{N}(x_t; \sqrt{1-\beta_t}x_{t-1}, \beta_t I).
\end{equation}
Here $\beta_t \in (0,1)$ is a predefined noise schedule parameter controlling the noise strength added at each step. Through the reparameterization trick \cite{kingma2022autoencodingvariationalbayes}, we can directly sample $x_t$ at any time $t$ from $x_0$ without iterative steps. Defining $\alpha_t := 1 - \beta_t$ and $\bar{\alpha}_t := \prod_{s=1}^{t} \alpha_s$, we have
\begin{equation}
q(x_t|x_0) = \mathcal{N}(x_t; \sqrt{\bar{\alpha}_t}x_0, (1-\bar{\alpha}_t)I),
\end{equation}
i.e.,
\begin{equation}
x_t = \sqrt{\bar{\alpha}_t}x_0 + \sqrt{1-\bar{\alpha}_t}\epsilon,
\end{equation}
where $\epsilon \sim \mathcal{N}(0,I)$. When $T$ is sufficiently large and the noise schedule is reasonable, $\bar{\alpha}_T \approx 0$, making $q(x_T|x_0) \approx \mathcal{N}(0,I)$, i.e., the final state approximates a Gaussian distribution \cite{ho2020ddpm}.

The reverse process aims to learn the temporal inversion of the forward process, i.e., starting from the noise distribution $p(x_T) = \mathcal{N}(0,I)$ and progressively denoising to recover the data distribution. The reverse process is also modeled as a Markov chain
\begin{equation}
p_\theta(x_{0:T}) := p(x_T) \prod_{t=1}^{T} p_\theta(x_{t-1}|x_t),
\end{equation}
where each reverse transition kernel is parameterized as
\begin{equation}
p_\theta(x_{t-1}|x_t) = \mathcal{N}(x_{t-1}; \mu_\theta(x_t, t), \Sigma_\theta(x_t, t)).
\end{equation}
In practice, the covariance matrix $\Sigma_\theta(x_t, t)$ is set to fixed values or time-dependent functions \cite{nichol2021iddpm}, while the mean $\mu_\theta(x_t, t)$ is modeled through neural networks. A key insight of DDPM~\cite{ho2020ddpm} is reparameterizing the mean as noise prediction:
\begin{equation}
\mu_\theta(x_t, t) = \frac{1}{\sqrt{\alpha_t}}\left(x_t - \frac{\beta_t}{\sqrt{1-\bar{\alpha}_t}}\epsilon_\theta(x_t, t)\right),
\end{equation}
where $\epsilon_\theta(x_t, t)$ is a neural network that predicts the noise $\epsilon$ added to $x_0$ at time $t$. This parameterization transforms the complex distribution modeling problem into a relatively simple noise regression problem \cite{ho2020ddpm}.

Based on the above definitions of forward and reverse processes, the training objective of diffusion models can be clearly constructed. The core is to minimize the KL divergence between the reverse process and the true posterior distribution. Through variational lower bound (ELBO) \cite{kingma2022autoencodingvariationalbayes} derivation, the final training loss can be simplified to
\begin{equation}
L = \mathbb{E}_{t \sim U[1,T], x_0 \sim q(x_0), \epsilon \sim \mathcal{N}(0,I)} \left[ \lambda(t) \|\epsilon - \epsilon_\theta(x_t, t)\|^2 \right],
\end{equation}
where $\lambda(t)$ is a weighting function and $x_t$ is computed from $x_0$ and $\epsilon$ through the reparameterization trick. The intuitive meaning of this loss function is to train the network to predict the noise added to clean images at given noise levels. The specific training process includes sampling clean images $x_0$ from the dataset, randomly sampling time step $t$ and noise $\epsilon$, computing $x_t$ through the forward process, training the network $\epsilon_\theta$ to predict noise $\epsilon$, and finally computing prediction error and backpropagating to update parameters \cite{ho2020ddpm}.

After training, the process of generating new samples is sampling from the reverse Markov chain. First, sample from the prior distribution $x_T \sim \mathcal{N}(0,I)$, then iteratively denoise for $t = T, T-1, \ldots, 1$: use the trained network to predict noise $\hat{\epsilon} = \epsilon_\theta(x_t, t)$, compute the denoised mean
\begin{equation}
\mu = \frac{1}{\sqrt{\alpha_t}}\left(x_t - \frac{\beta_t}{\sqrt{1-\bar{\alpha}_t}}\hat{\epsilon}\right),
\end{equation}
sample the next state $x_{t-1} = \mu + \sigma_t z$ (where $z \sim \mathcal{N}(0,I)$), and finally return the generated sample $x_0$. Standard DDPM requires complete $T$-step iteration (usually $T=1000$), leading to slow inference speed. Subsequent works such as DDIM\cite{song2021ddim} and DPM-Solver\cite{lu2022dpmsolver} reduce the number of steps to dozens through deterministic sampling or high-order numerical methods, significantly improving speed while maintaining quality.

From a broader theoretical perspective, beyond the DDPM viewpoint, the theoretical framework of diffusion models can also be unified and generalized through Score Matching and Stochastic Differential Equations (SDEs)\cite{song2021sde}. Although these perspectives differ in mathematical form, they ultimately all point to the core idea of learning the data distribution gradient (the score function) through neural networks, and provide a theoretical basis for the design of more flexible samplers, such as ODE solvers.

In recent years, Flow Matching\cite{lipman2023flowmatchinggenerativemodeling} has emerged as a further theoretical development of diffusion models, providing a more concise and efficient training paradigm building upon DDPM. Unlike traditional DDPM which relies on stochastic diffusion processes and complex variational lower bound derivations, Flow Matching directly learns deterministic Continuous Normalizing Flows (CNF) \cite{chen2019neuralordinarydifferentialequations} from noise distribution to data distribution. Specifically, given a target probability path $p_t(x)$ and its corresponding generating vector field $u_t(x)$, Flow Matching trains a neural network $v_\theta$ to regress the target velocity field:
\begin{equation}
L_{FM}(\theta) = \mathbb{E}_{t, p_t(x)} \left[ \| v_\theta(x, t) - u_t(x) \|^2 \right],
\end{equation}
where $t \sim U[0,1]$. Since the global probability path $p_t(x)$ is difficult to construct directly, Conditional Flow Matching\cite{lipman2023flowmatchinggenerativemodeling} only requires defining conditional probability paths $p_t(x|x_1)$ and corresponding conditional velocity fields $u_t(x|x_1)$ at the sample level, thus avoiding explicit modeling of global paths and greatly simplifying the training process. Rectified Flow\cite{liu2022flowstraightfastlearning}, as an important variant of Flow Matching, learns straight-line interpolation paths from data $x_0$ to noise $x_1$, reducing the number of ODE solver steps and further improving sampling efficiency. Flow Matching has gradually become a mainstream approach in the diffusion model field and has been widely adopted in large-scale generative models including Stable Diffusion 3\cite{esser2024scalingrectifiedflowtransformers} and FLUX\cite{blackforestlabs2024flux}, providing a more flexible theoretical foundation for subsequent model architecture design and caching optimization techniques.

From a computational perspective, the Iterative Inference Mechanism of diffusion models is inherently a highly redundant process. Due to the continuous and progressive nature of denoising steps in the temporal dimension, feature representations between adjacent time steps tend to be highly similar, and the model's intermediate layer activations and attention structures also exhibit significant temporal correlations. This computational overhead means that there are many reusable intermediate results in the multi-step denoising process. Based on this characteristic, researchers have proposed the Diffusion Caching mechanism. By caching and reusing computational results with high correlation between adjacent time steps, it effectively reduces unnecessary calculations without changing the model structure and parameters, thereby significantly improving inference efficiency and scalability.

\subsection{Caching Acceleration Techniques}

\subsubsection{U-Net Architecture}
The U-Net architecture adopts a symmetric encoder-decoder design, capturing contextual information through a contracting path and achieving precise localization through an expanding path \cite{ronneberger2015unetconvolutionalnetworksbiomedical}. Given noisy input $x_t \in \mathbb{R}^{H \times W \times C}$, U-Net consists of three core components: \textbf{encoder}, \textbf{bottleneck layer}, and \textbf{decoder}. The encoder extracts hierarchical feature pyramids $\{h_1^{enc}, h_2^{enc}, \ldots, h_L^{enc}\}$ through consecutive downsampling operations, while the decoder reconstructs output through skip connection mechanisms, concatenating encoder features with upsampled features, where $\text{Concat}(\cdot, \cdot)$ represents the feature concatenation operation.

The caching opportunities in U-Net arise from the evolutionary differences of features across layers in the temporal dimension. Empirical studies \cite{ma2023deepcacheacceleratingdiffusionmodels} reveal that \textbf{high-level features} $h_L^{enc}$ encode global semantic information and exhibit significant temporal consistency between adjacent denoising steps, while \textbf{low-level features} $h_1^{enc}$ are responsible for detail texture reconstruction and demonstrate higher sensitivity to time step changes. This evolutionary difference across layers provides theoretical justification for designing layer-selective caching strategies. Slowly-changing high-level semantic features can be cached and reused while maintaining dynamic updates of low-level texture features, thereby significantly reducing computational overhead while preserving generation quality.

\subsubsection{Transformer Architecture}
Diffusion Transformers (DiT)\cite{peebles2023scalablediffusionmodelstransformers} shift the image processing paradigm from convolution operations to  attention mechanisms, adopting Vision Transformer \cite{dosovitskiy2020image} design principles. Given latent representation $z_t$, DiT first decomposes it into non-overlapping patch sequences
\begin{equation}
x_t = \text{Patchify}(z_t) \in \mathbb{R}^{N_p \times d},
\end{equation}
where $N_p = (H/p) \times (W/p)$ is the number of patches and $p$ is the patch size. The overall architecture consists of $L$ DiT blocks in sequence (where $L$ denotes the total number of layers), with each DiT block $g_l$ ($l \in \{1, 2, \ldots, L\}$) adopting Transformer design but injecting time step and class conditions through Adaptive Layer Normalization (AdaLN) mechanisms:
\begin{equation}
g_l = \text{AdaLN}(\text{MLP}(\text{AdaLN}(\text{Attention}(x), t, c)), t, c),
\end{equation}
where
\begin{equation}
\text{AdaLN}(x, t, c) = \gamma(t, c) \cdot \text{LN}(x) + \beta(t, c),
\end{equation}
and parameters $\gamma, \beta$ are computed from time step $t$ and condition $c$ through MLP networks \cite{peebles2023scalablediffusionmodelstransformers,vaswani2023attentionneed}.

Unlike U-Net's spatial hierarchy, DiT's optimization space is primarily manifested in two aspects. First, the temporal stability of attention patterns, where token-wise attention weight matrices $A_{ij} = \text{softmax}(QK^T)$ exhibit relatively stable patterns during specific stages of the denoising process, providing possibilities for caching attention results. Second, each DiT block $g_l$ typically comprises multiple computational modules including self-attention $F^l_{SA}$, cross-attention $F^l_{CA}$ (when applicable for conditional generation), and feed-forward network $F^l_{MLP}$. These components demonstrate different sensitivities to timesteps, with $F^l_{CA}$ having higher caching potential due to its dependence on relatively stable conditional information.

To establish a unified analytical framework, this subsection formally defines caching acceleration techniques. The essence of caching acceleration techniques lies in identifying and exploiting computational redundancy in inference processes.

The caching acceleration techniques in diffusion models share similar design principles with the KV-Cache mechanism in large language models (LLMs). In LLM autoregressive generation, the Key and Value matrices of already-generated tokens are fixed and unchanging, and KV-Cache avoids redundant computation on historical sequences by caching and reusing these matrices. Diffusion model caching strategies similarly aim to reduce redundant computation through reusing historical computation results.

 Specifically, we define the caching operation for any computational module (whether U-Net’s convolutional layers or DiT’s Attention and MLP blocks) as follows. 
Let $N$ denote the \textit{cache reuse interval}, indicating that cached features computed at step $t-N$ are reused for the subsequent $N-1$ timesteps. 

For the $l$-th layer of the network, the caching process first computes and stores the output at timestep $t$:
\begin{equation}
\mathcal{C}^l_t := F^l(x_t),
\end{equation}
where $F^l$ denotes the $l$-th layer’s forward function and $x_t$ is the input at time $t$. 
Within the following $N-1$ steps, these cached features are directly reused for intermediate timesteps $\{t-1, t-2, \ldots, t-(N-1)\}$:
\begin{equation}
F^l(x_{t-k}) \approx \mathcal{C}^l_t, \quad k \in \{1, 2, \ldots, N-1\}.
\end{equation}
This strategy effectively skips redundant forward computations for these $N-1$ steps, yielding an $(N-1)$-fold computational acceleration within each cache reuse interval.

The acceleration effect of caching strategies stems from significant reduction in computational complexity. Based on standard algorithmic complexity theory \cite{cormen01introduction}, consider the complete inference process with $T$ total denoising steps, where a caching strategy performs full computation at $m$ steps while the remaining $(T-m)$ steps reuse cached features. This reduces the total computational complexity from $\mathcal{O}(T \cdot C_1)$ to $\mathcal{O}(m \cdot C_1 + (T-m) \cdot C_2)$, where $C_1$ denotes the complexity of full forward computation and $C_2$ denotes the complexity of cache retrieval. Since cache access is typically orders of magnitude faster than full computation (i.e., $C_2 \ll C_1$), this yields an acceleration factor of approximately $T/m$.

However, naive caching strategies face fundamental challenges when dealing with the dynamic evolution of features. The denoising process in diffusion models is inherently dynamic, with feature representations continuously changing along the temporal dimension—\textit{remaining relatively stable over short intervals but exhibiting significant drift as time accumulates}. Directly reusing historical features therefore leads to cumulative errors over multiple inference steps. This characteristic necessitates that caching strategies strike a balance between acceleration gains and error control, which constitutes the core design objective of subsequent caching methods.

\subsection{Static Caching Methods}

\textbf{\textit{Static Caching}} \textit{uses a fixed reuse strategy in which caching is performed at predefined layers or timesteps and remains constant across all inference runs, independent of content or input}. This approach is simple to implement and highly stable, but it lacks flexibility as it does not dynamically adjust according to content changes during the process.


As one of the earliest works to introduce caching mechanisms into diffusion models, \textit{DeepCache}~\cite{ma2023deepcacheacceleratingdiffusionmodels} utilizes the observation of the U-Net structure. It takes advantage of the fact that the Upsampling layer features change little during the continuous denoising process, and it preserves the computations for the Downsampling layers while directly reusing the upsampling features. \textit{DeepCache} adopts a fixed-interval strategy, performing a complete computation of the entire U-Net at a specific time step \(t\). During this computation, it caches the upsampling layer features \(U_{m+1}^t(\cdot) \) generated by the main branch, specifically caching the feature results after the \(m+1\)-th upsampling block layer:
\begin{equation}
F_{\text {cache }}^t \leftarrow U_{m+1}^t(\cdot)
\end{equation}

In the subsequent time steps, the newly computed downsampling layer features \( D_m^{t-1}(\cdot) \) are concatenated with the cached upsampling layer features \( F_{\text{cache}}^t \) retrieved from the cache, and used as the input for the \( m \)-th upsampling block:

\begin{equation}
U^{t-1}_{m}\leftarrow \operatorname{Concat}\left(D_m^{t-1}(\cdot), F_{\text{cache}}^t\right)
\end{equation}

\textit{DeepCache} leverages the ``laziness'' of the upsampling layer feature changes, using caching to avoid redundant computations. At the same time, it retains the frequent updates of the downsampling layer features to ensure that the quality and details of the generated images are progressively improved.

\textit{FasterDiffusion}~\cite{li2024faster} quantitatively measures the inter-step feature differences across layers of a U-Net based model and observes that the encoder features exhibit significantly smaller magnitude and variance of change compared to those in the decoder. This finding reveals the stability of encoder representations and the dynamic nature of decoder features. Based on this core insight, \textit{FasterDiffusion} introduces Encoder Propagation. Specifically, the diffusion process is divided into \emph{key timesteps} and \emph{non-key timesteps}. At a key timestep $t$, the model performs a full forward pass and stores the multi-level outputs of the encoder $E$. During the subsequent $K$ non-key timesteps $t-1, t-2, \ldots, t-k+1$, the model completely skips the computation of the encoder $E$, while the decoder $D$ reuses the cached encoder features from timestep $t$. Since multiple decoder steps share identical encoder inputs, their computations can be executed in parallel, thereby breaking the traditional sequential constraint and achieving substantial inference acceleration.

To compensate for the slight texture detail loss potentially caused by encoder propagation, \textit{FasterDiffusion} further introduces a lightweight mechanism called {Prior Noise Injection}. In the late stage of generation ($t < \tau$), a small proportion $\alpha$ of the initial noise latent $z_T$ is injected into the current latent variable $z_t$, i.e.,
\begin{equation}
z_t = z_t + \alpha \cdot z_T.
\end{equation}
This simple operation effectively enhances the high-frequency details of generated images while incurring almost no additional computational cost
\cite{li2024faster}.


Unlike the U-Net architecture, the DiT architecture does not have the structural characteristics of upsampling and downsampling layers, which makes the \textit{DeepCache} method based on structural features inapplicable to the DiT-based generative models. Methods represented by \textit{PAB}~\cite{zhao2024real} focus on the trend of changes in the attention mechanism during the denoising process. They find that different types of attention outputs have different redundancies. Specifically, spatial attention shows the largest variation, involving high-frequency elements such as edges and textures; temporal attention exhibits mid-frequency changes related to motion and dynamics in videos; and cross-modal attention is the most stable, linking text with video content, similar to low-frequency signals that reflect textual semantics. Based on the redundancy of different types of attention, \textit{PAB} designs the \textbf{P}yramid \textbf{A}ttention \textbf{B}roadcast to reduce unnecessary attention computations. It customizes different broadcast ranges for each attention type and MLP layers based on their stability and output variation rate. Attention types with more changes and fluctuations are assigned smaller broadcast ranges, while those with fewer changes are given larger ranges, forming a ``pyramid'' strategy.


\textit{FORA}~\cite{selvaraju2024fora} provides a more comprehensive analysis of the diffusion process, finding that the output images of consecutive time steps during sampling exhibit significant visual similarity, which is highly correlated with the similarity of hierarchical features. Based on the constancy of the Diffusion model structure throughout the entire sampling process, \textit{FORA} addresses the major computational cost of the DiT model by introducing a universal caching mechanism for the Self-Attention and MLP layers. With a fixed interval \(N\), after a full computation, all the corresponding output features of the model layers are cached. Initially, the model performs a full computation for all layers of the DiT block. For each layer \(k\), the computed Attention Features are stored in \( F_{attn}^{k} \), and the computed MLP Features are stored in \( F_{mlp}^{k} \). For the next \(N-1\) time steps, until the next cycle begins, the model reuses the cached features instead of recomputing them. This means that during these steps, for any layer \(k\), the model retrieves features from \(F_{attn}^{k}\) and \(F_{mlp}^{k}\), avoiding the computational cost of full computation for each step. \textit{FORA} is a simple and effective cache mechanism proposed for any DiT-based model. It provides a static caching paradigm for DiT-based models and has inspired subsequent improvements in this area.


Considering the differences among different DiT layers, \textit{\(\Delta\)-DiT}~\cite{chen2024delta} delves deeper into the correlation between different layers of DiT and image generation. The study finds that the front layers of DiT are related to the overall contours of the generated image, while the rear layers are related to the details. Based on this insight, \textit{\(\Delta\)-DiT} caches the rear layers during the early generation stages (contour phase) and caches the front layers during the later stages (detail phase), achieving adaptive acceleration that aligns with the generation process. 

Unlike the residual form of U-Net, directly caching the output feature maps of DiT layers would completely lose the information from the previous sampling step \(x_{t-1}\) that the block depends on. To address the information loss caused by direct caching, \textit{\(\Delta\)-DiT} propose \(\Delta\)-Cache, which caches the ``increment'' (residual) of the features instead of their absolute values. Specifically, for \(k\)-th layer of DiT, \(\Delta\)-Cache caches \( F^{k}(x_t) - x_t \). In the next step \( t-1 \), the approximation calculation can be represented as \( x_{t-1} + (F^{k}(x_t) - x_t) \). This method incorporates the previous sampling result \( x_{t-1} \) into the calculation, avoiding information loss. \textit{\(\Delta\)-DiT} is particularly suitable for DiT’s isotropic architecture, as the input and output feature maps of each block are of consistent scale, making it easier to compute the difference. However, \textit{\(\Delta\)-DiT} also has limitations, as the caching mechanism is overly aggressive, skipping a large number of DiT layers at once. This high acceleration ratio can lead to content degradation.


\textit{FasterCache}~\cite{lv2024fastercache} further optimizes time redundancy introduced by the unconditional output of CFG calculations. It utilizes a weight function \( w(t) \) that linearly increases from 0 to 1 with each time step \( t \), to blend the cached features (such as \( F_{\text{cache}}^t \) and \( F_{\text{cache}}^{t+2} \)), thereby preserving subtle changes during the denoising process. Additionally, \textit{CFG-Cache} is introduced, which leverages frequency domain decomposition. It uses high-pass (HPF) and low-pass (LPF) filters to cache the difference between conditional and unconditional outputs. In subsequent inference steps, the unconditional output can be efficiently computed using the cached frequency domain information and adaptive weights \( w_1(t) \) and \( w_2(t) \), avoiding the need for full unconditional forward propagation.

The recently proposed \textit{FORA} and \(\Delta\)-DiT directly applied prior caching methods to DiT, but they did not fully analyze or leverage the specific characteristics of the Transformer architecture. \textit{ToCa}~\cite{zou2024accelerating} first investigates how caching techniques impact DiT in the token level, revealing that different tokens exhibit varying sensitivity to feature caching. This difference arises from two main factors: first, different tokens have different redundancies in the time dimension; second, the same caching error on different tokens can lead to significantly different biases in the final generated output. \textit{ToCa} provides a fine-grained caching strategy for tokens within the same layer and across different layers, allowing for the accurate selection of tokens suitable for feature caching with minimal computational cost. 

Furthermore, \textit{ToCa}~\cite{zou2024accelerating} defines four scoring metrics for token selection from two perspectives: \textit{Temporal Redundancy }and \textit{Error Propagation}. These scores can be obtained without introducing additional computational overhead. The caching score for each token is defined as:

\begin{equation}
S\left(x_i\right)=\sum_{j=1}^4 \lambda_j s_j\left(x_i\right)
\end{equation}

The tokens with the lowest scores are selected as caching candidates:

\begin{equation}
\begin{aligned}
    \mathcal{I}_{\text{Cache}} = & \underset{\substack{
        \left\{i_1, i_2, \ldots, i_{R \% \times N}\right\} \\
        \subseteq \{1, 2, \ldots, n\}
    }}{\arg \min} \left\{ \mathcal{S}\left(x_{1}\right), \mathcal{S}\left(x_{2}\right), \ldots, \mathcal{S}\left(x_{n}\right) \right\}
\end{aligned}
\end{equation}

In \textit{ToCa}, the overall caching ratio is first set as a global hyperparameter \( R_0 \), which controls the allowable caching scale during inference. However, a single global ratio cannot adequately address the varying needs across different layers, token types, and time steps. To this end, \textit{ToCa} introduces a multidimensional dynamic adjustment mechanism based on the global ratio, where the specific caching ratio for the \(l\)-th layer, token type \textit{type}, and time step \(t\) is defined as:

\begin{equation}
R_{l,t,\text{type}} = R_0 \times r_l \times r_{\text{type}} \times r_t
\end{equation}

By leveraging the differences between tokens, \textit{ToCa} achieves significantly better acceleration performance in image and video generation compared to existing caching methods, with almost no loss in generation quality. It provides a new approach for token-wise optimization in the DiT-based model. 

\subsection{Dynamic Caching Methods}

Unlike the fixed strategy of Static Caching, \textbf{\textit{Dynamic Caching}} introduces an error checking mechanism, representing the evolution of caching technology towards refinement. \textit{At each layer or step of inference, it dynamically decides whether to perform computation, update the cache, or directly use the cache based on predefined metrics }(such as feature similarity, $L1$-Norm, etc.). This adaptability enables ``\textit{on-demand computation}'' and achieves a higher acceleration ratio. The adaptive adjustment of \textbf{\textit{Dynamic Caching}} can be expanded along multiple dimensions, which will be introduced separately below.

\subsubsection{Timestep-Adaptive Caching Methods}
Diffusion Models achieve high-fidelity generation through a gradual denoising process, yet this multi-step inference inherently involves substantial temporal redundancy. Early caching methods (e.g., \textit{DeepCache} and \textit{FORA}) typically adopt \textbf{\textit{fixed-interval static strategies}}, where cached features from one forward pass are reused for $k$ consecutive timesteps. Although such strategies reduce computation to some extent, they implicitly assume that feature variations are uniformly distributed across timesteps. However, empirical studies reveal that the evolution of features in the diffusion process is highly non-uniform: \textit{features vary smoothly and are highly reusable during early stages, while rapid changes occur in later stages, leading to reduced correlation}. Consequently, fixed-interval reuse often results in accuracy degradation and generation quality loss.  

\textbf{\textit{Timestep-Adaptive Caching methods}} dynamically adjust caching activation and refresh timing based on the varying stability and dynamics of features across diffusion stages. By estimating model outputs, internal feature dynamics, or temporal stability online, these approaches flexibly determine \emph{when to compute and when to reuse}, significantly reducing redundant computation while maintaining visual fidelity.

As representative method, \textit{TeaCache}~\cite{liu2025timestep} introduces an input-side signal-based dynamic change estimation mechanism. It computes the difference between noise-conditioned features modulated by timestep embeddings to predict output variations between adjacent timesteps, thereby deciding when to refresh cached features. \textit{TeaCache} defines the relative $L1$ difference between adjacent outputs $O_t$ and $O_{t+1}$ as:
\begin{equation}
L1_{\text{rel}}(O,t) = \frac{\|O_t - O_{t+1}\|_1}{\|O_t\|_1 + \|O_{t+1}\|_1}.
\end{equation}
Since input-level differences and actual output variations differ in scale, polynomial fitting is used for correction:
\begin{equation}
\hat{y} = a_0 + a_1x + a_2x^2 + \cdots + a_nx^n,
\end{equation}
where $a_i$ are the fitted coefficients. During inference, the model accumulates corrected estimates and performs a full computation only when the cumulative variation exceeds a threshold $\delta$:
\begin{equation}
\sum_{t=t_a}^{t_b-1}\hat{L1}_{\text{rel}}(O,t) < \delta.
\end{equation}
This \emph{change-driven refresh} paradigm allows \textit{TeaCache} to substantially reduce redundant computation while maintaining visual fidelity. Owing to its excellent trade-off between efficiency and generation quality, \textit{TeaCache} has become one of the most influential and widely adopted caching methods in diffusion acceleration research and practical systems.

Another research direction recognizes that different diffusion stages exhibit distinct computational demands. \textit{VCUT}~\cite{taghipour2025faster} systematically analyzes the role of cross-attention in Stable Video Diffusion and finds that due to the global pooling property of CLIP embeddings, cross-attention contributes minimally—particularly in mid-to-late stages. Based on this, \textit{VCUT} divides the diffusion process into a \emph{semantic binding stage} and a \emph{quality refinement stage}: the former ensures semantic consistency, while the latter focuses on detail enhancement. During the semantic binding stage, \textit{VCUT} computes conditional linear outputs as:
\begin{equation}
M = \frac{1}{2}\big(L_{\tau(y)}^{c,m} + L_{\varnothing}^{c,m}\big), \quad m \in [1, l],
\end{equation}
and reuses these cached features in later stages, effectively removing redundant cross-attention computation. This stage-based reuse strategy achieves substantial computational savings without compromising semantic consistency.

In contrast to heuristic or stage-based estimation, \textit{LazyDiT}~\cite{shen2025lazydit} explicitly learns when to skip computations across timesteps. Its key insight is that, during diffusion inference, certain modules (e.g., attention or feed-forward layers) produce nearly identical outputs between adjacent timesteps and can thus be reused. \textit{LazyDiT} introduces a linear predictor before each Transformer layer to learn a similarity function based on a first-order Taylor approximation:
\begin{equation}
f(Y^{\Phi}_{\ell, t-1}, Y^{\Phi}_{\ell, t}) \approx \langle W^{\Phi}_{\ell}, Z^{\Phi}_{\ell,t} \rangle.
\end{equation}
If the predicted similarity exceeds a threshold, the computation for that layer is skipped and the cached output is reused. During training, a \emph{lazy loss} is introduced to encourage the model to learn optimal skipping behavior:
\begin{equation}
\mathcal{L}_{\text{lazy},t} =
\rho_{\text{attn}}\frac{1}{B}\sum_{l=1}^L\sum_{b=1}^B(1-s^{\text{attn}}_{l,t})_b +
\rho_{\text{feed}}\frac{1}{B}\sum_{l=1}^L\sum_{b=1}^B(1-s^{\text{feed}}_{l,t})_b.
\end{equation}
Experiments demonstrate that \textit{LazyDiT} maintains comparable generation quality even when skipping approximately 50\% of layer computations, offering a fine-grained and learnable framework for timestep-adaptive acceleration.

At a deeper level of temporal decomposition, the \textit{T-GATE} series reveals distinct functional roles of self-attention and cross-attention across diffusion stages. Studies show that cross-attention dominates semantic planning in early stages but contributes little later, while self-attention becomes critical for fine-grained detail synthesis. Accordingly, \textit{T-GATE V1}~\cite{zhang2024cross} caches cross-attention outputs after the semantic stage:
\begin{equation}
F = \Big\{ \tfrac{1}{2}(C_{\emptyset}^{m,i} + C_c^{m,i}) \,\Big|\, i \in [1, l] \Big\},
\end{equation}
and reuses them in subsequent stages. \textit{T-GATE V2}~\cite{liu2024faster} further introduces a phase-based policy: adopting a \textit{``warm-up–interval reuse'' }schedule for attention, skipping computations in early steps and performing full calculations later,  achieving a balanced trade-off between speed and fidelity.

\textit{Chipmunk}~\cite{silveria2025chipmunk} redefines the granularity of timestep redundancy by extending caching to the activation level. By observe that $5\%–25\%$ of activations account for $70\%–90\%$ of total variation across timesteps, they propose a sparse incremental computation framework: performing full computation (\textit{``dense steps''}) intermittently and caching key indices, while in \textit{``sparse steps''} recomputing only high-contribution activations and reusing the rest. With a hardware-friendly column-sparse implementation, \textit{Chipmunk} significantly accelerates inference while preserving generation quality.

Transitioning from empirical fitting to physical modeling, \textit{MagCache}~\cite{ma2025magcache} investigates residual evolution dynamics and proposes a \emph{unified amplitude decay law}. It shows that the magnitude ratio of adjacent residuals
\begin{equation}
r_t = v_\theta(x_t, t) - x_t, \quad 
\gamma_t = \frac{\|r_t\|_2}{\|r_{t-1}\|_2}
\end{equation}
monotonically decreases over time, with nearly constant directional consistency (token-level cosine distance $\approx 0$). Thus, skip-step error can be modeled as a geometric decay:
\begin{equation}
\varepsilon_{\text{skip}}(\hat{t}, t) = 1 - \prod_{i=\hat{t}+1}^{t}\gamma_i.
\end{equation}
By maintaining an accumulated error $E_t$ and refreshing when it exceeds a threshold, \textit{MagCache} achieves robust performance across models and prompts without offline fitting.

Finally, \textit{EasyCache}~\cite{zhou2025less} emphasizes a fully online self-correction mechanism. Analysis of DiT inference trajectories reveals a nearly constant \emph{relative transformation rate}:
\begin{equation}
k_t = \frac{\|v_t - v_{t-1}\|}{\|x_t - x_{t-1}\|},
\end{equation}
indicating strong local linearity in diffusion processes. Hence, the transformation vector from the last full computation $\Delta_i = v_i - x_i$ can approximate future outputs:
\begin{equation}
\hat{v}_t = x_t + \Delta_i.
\end{equation}
An accumulated deviation indicator
\begin{equation}
E_t = \sum_{n=i+1}^{t}\varepsilon_n, \quad 
\varepsilon_n \approx \frac{k_i\|x_n - x_{n-1}\|}{\|v_{n-1}\|} \times 100\%
\end{equation}
is used to monitor caching error in real time. The cache is refreshed only when $E_t \ge \tau$, forming a purely online gating mechanism free from offline calibration. \textit{EasyCache} dynamically adapts computational intensity according to model state, maintaining a stable balance between inference efficiency and generative quality.

\subsubsection{Layer-Adaptive Caching Methods}
In diffusion models, different network layers exhibit distinct temporal behaviors during the generation process. 
Shallow layers primarily capture low-level textures and local spatial details, showing relatively smooth and stable variations across timesteps. 
In contrast, deeper layers encode high-level semantics and global structures, whose representations evolve more rapidly and irregularly over time. 
Such layer-wise diversity in temporal dynamics results in non-uniform feature stability across the network, 
implying that a single, fixed caching policy cannot optimally accommodate all layers.

To address this issue, \textbf{\textit{Layer-Adaptive Caching Methods}} aim to determine both \textit{where to compute and where to reuse} within the model, 
adaptively adjusting the caching and update frequency of each layer according to its dynamic characteristics—such as gradient magnitude, feature difference, or similarity metrics. 
Compared with timestep-adaptive approaches, layer-adaptive caching focuses on the structural heterogeneity within the network, 
providing fine-grained control over computational allocation and achieving efficient acceleration while maintaining generation stability and fidelity.

\textit{Cache Me if You Can}~\cite{wimbauer2024cache} first proposed Block Caching, whose core idea is to exploit the redundancy of internal computations within the denoising network. Instead of treating the denoising U-Net as a black box, the researchers analyzed the behavioral changes of its internal blocks during iterative inference. They found that the outputs of U-Net blocks vary smoothly across timesteps, with different patterns across layers, but with overall minimal inter-step differences. They defined the relative absolute variation between blocks as:
\begin{equation}
L1_{\text{rel}}(i, t) = \frac{\|C_i(x_t, s_t) - C_i(x_{t-1}, s_{t-1})\|_1}{\|C_i(x_t, s_t)\|_1},
\end{equation}
where \(C_i\) denotes the core computation output of the \(i\)-th block. Based on this metric, Block Caching caches the block output computed at timestep \(t_a\) and reuses it until the cumulative variation exceeds a threshold \(\delta\):
\begin{equation}
\sum_{t=t_a}^{t_b-1} L1_{\text{rel}}(i,t) \le \delta < \sum_{t=t_a}^{t_b} L1_{\text{rel}}(i,t).
\end{equation}
This method achieves fine-grained block-level scheduling: computationally expensive blocks (e.g., SpatialTransformer) are cached when variation is small, while detail-sensitive blocks (e.g., ResBlock) are refreshed more frequently. To alleviate feature misalignment artifacts, the authors introduced a lightweight \emph{scale-shift adjustment} mechanism, applying timestep-dependent scaling and shifting to cached features, and employing a student–teacher distillation process to ensure feature distribution consistency. Overall, Block Caching significantly accelerates inference while maintaining or even improving visual quality.

However, non-learning caching methods that rely on fixed rules or thresholds still suffer from distribution mismatch between training and inference. To mitigate this limitation, \textit{HarmoniCa}~\cite{huang2024harmonica} reconceptualizes learning-based caching from a system-level perspective, enabling adaptive and efficient feature reuse. The method pointed out that traditional Learning-to-Cache frameworks optimize only randomly sampled single timesteps during training, preventing the model from learning the true temporal dependencies and error accumulation patterns observed during inference. To overcome this, \textit{HarmoniCa} introduced \emph{Stepwise Denoising Training (SDT)}, which performs full denoising trajectories from Gaussian noise \(x_T\) to the final image \(x_0\) during training, enabling the student model to perceive the historical influence of caching decisions under teacher guidance. Furthermore, it proposed the \emph{Image Error Proxy Objective} (IEPO) as follows:
\begin{equation}
\mathcal{L}_{\text{IEPO}}^{(t)} = \lambda^{(t)} \mathcal{L}_{\text{MSE}}^{(t)} + \beta \sum_{i=0}^{N-1} r_{t,i},
\end{equation}
where \(\lambda^{(t)} = \|x_0 - x_0^{(t)}\|_F^2\) dynamically reflects the quality gap between generated and target images, \(\mathcal{L}_{\text{MSE}}^{(t)}\) maintains teacher–student consistency, and the regularization term \(\sum r_{t,i}\) encourages the model to reuse cached results through learnable caching weights. This objective optimizes for final image quality and eliminates the training–inference distribution gap.

For the more complex scenario of video generation, \textit{AdaCache}~\cite{kahatapitiya2024adaptive} further extended caching with spatiotemporal adaptivity. Video diffusion models must process both inter-frame temporal continuity and large-scale spatial features, with significant motion variations across regions. \textit{AdaCache} reformulates caching as a content-adaptive dynamic scheduling problem. Specifically, it measures the residual variation:
\begin{equation}
c_t^l = \|r_t^l - r_{t-k}^l\|_1,
\end{equation}
to quantify changes in the residuals of layer \(l\) across diffusion steps, and introduces a motion regularization term
\begin{equation}
mg_t^l = \frac{\partial m_t^l}{\partial t},
\end{equation}
to characterize motion intensity in latent space. These are combined into a composite metric \(\tilde{c}_t^l = \alpha c_t^l + \beta mg_t^l\), enabling aggressive caching in static regions and frequent updates in dynamic ones. This strategy achieves content-aware hierarchical scheduling for video diffusion, significantly accelerating inference while maintaining high generation quality.

\textit{FEB-Cache}~\cite{zou5584552feb} develops a frequency-domain-driven caching mechanism to mitigate exposure bias.
It shows that Attention and MLP layers exhibit complementary spectral sensitivities: \textit{Attention emphasizes low-frequency structures, while MLP captures high-frequency details}, making unified caching suboptimal and prone to error amplification. \textit{FEB-Cache} introduced dynamic noise scaling:
\begin{equation}
\varepsilon_t \rightarrow 
\begin{cases}
a \cdot \varepsilon_t, & t > T_0,\\
b \cdot \varepsilon_t, & t \le T_0,
\end{cases}
\end{equation}
and constructed a frequency-oriented cache table that prioritizes caching MLP modules in early stages and Attention modules in later stages to align with the stage-specific frequency composition of exposure bias. By comparing the error of three caching states: \textit{NoCache, AllCache, and StageSpecific}, the optimal strategy is selected to dynamically balance acceleration and quality across stages.

In contrast to methods requiring retraining or frequency statistics, \textit{DBCache}~\cite{cache-dit@2025} proposed a training-free \emph{Dual Block Caching} framework. Inspired by U-Net architecture design, \textit{DBCache} divides the Transformer block stack of DiT into three functional segments: a front section (controlled by \(F_n\)) acting as a \textit{``probe''} that performs full computation to capture residual signals for comparison with the previous step; a middle section serving as the main caching region, skipping computation and reusing cached outputs when residual changes fall below a threshold \(\text{residual\_diff\_threshold}\); and a rear section (controlled by \(B_n\)) functioning as a \textit{``corrector''} that always recomputes to fuse and correct possible deviations. This  \textit{``probe–decision–correction''} loop enables structured caching that can be flexibly applied to  DiT models.

\textit{Foresight}~\cite{adnan2025foresight} introduces a foresight-driven dynamic caching framework that leverages layer-wise feature heterogeneity in video diffusion models.
Since shallow layers exhibit high temporal similarity while deep layers vary more rapidly, Foresight estimates per-layer variation thresholds during a warm-up stage to guide adaptive caching throughout inference.
\begin{equation}
\lambda_l^x = \sum_{t=W-2}^{W} \frac{1}{10^{W-t}} 
\left(\frac{1}{P} \sum_{i=1}^{P} (x_l^i(t) - x_l^i(t-1))^2\right),
\end{equation}
where \(P\) denotes the number of feature elements. During inference, the reuse or recomputation of each layer is decided in real time based on the dynamic reuse metric \(\delta_l(t)\) and layer-specific threshold:
\begin{equation}
x_{l}^{t+1} =
\begin{cases}
C(x_l), & \text{if } \delta_l(t) \le \gamma \lambda_l,\\
\text{Compute}, & \text{otherwise}.
\end{cases}
\end{equation}

\subsubsection{Predictive Caching Methods} 
The iterative denoising process of diffusion models can be interpreted as a continuous dynamic system, 
where the hidden feature evolution over timesteps forms a trajectory that can be approximated by numerical prediction~\cite{liu2025reusing}. 
\textbf{\textit{Predictive Caching Methods}} leverage this insight by extending the traditional caching paradigm from simple reuse (\textbf{\textit{Cache-Then-Reuse}}) 
to forward prediction of future feature states (\textbf{\textit{Cache-Then-Forecast}}). 
Instead of relying solely on stored historical features, these methods leverage cached representations from previous timesteps to explicitly forecast future features over multiple steps, using numerical solvers or derivative-based predictors.

\textit{TaylorSeer}~\cite{liu2025reusing} serves as a pioneering work in the predictive caching domain, systematically challenging the theoretical foundation of the traditional \textit{``Cache-Then-Reuse''} paradigm. Existing cache-based acceleration methods for diffusion models commonly assume high similarity between adjacent timestep features, but this assumption often fails at high acceleration ratios, leading to rapid degradation of feature similarity as caching frequency increases. Through in-depth theoretical analysis and empirical studies, \textit{TaylorSeer} discovered a key insight: \textit{while the similarity of feature values themselves may decay, the evolutionary trajectory of features in the temporal dimension exhibits high regularity and predictability}. Through Principal Component Analysis (PCA), researchers observed that both features and their temporal derivatives (i.e., the ``velocity'' of feature changes along trajectories) present stable patterns. Based on this discovery, \textit{TaylorSeer} proposed the revolutionary \textit{``Cache-Then-Forecast''} paradigm, remodeling the problem as mathematical prediction of feature trajectories. 

The core idea of \textit{TaylorSeer} is to treat cached features as discrete samples along a continuous feature trajectory and model their evolution via Taylor series expansion. 
To estimate higher-order derivatives without additional computation, it employs finite difference approximations, allowing high-order feature dynamics to be inferred from a few fully computed timesteps. The prediction formula is:
\begin{equation}
    \mathcal{F}_{\textrm{pred},m}(x_{t-k}^l) = \mathcal{F}(x_t^l) + \sum_{i=1}^{m} \frac{\Delta^i\mathcal{F}(x_t^l)}{i! \cdot N^i}(-k)^i
\end{equation}
where $\Delta^i\mathcal{F}(x_t^l)$ is computed through finite difference methods, $N$ is the sampling interval, and $m$ is the order of Taylor expansion. The significance of \textit{TaylorSeer} lies not only in achieving the paradigm shift from passive reuse to active prediction, but more importantly in establishing the theoretical foundation for the entire predictive caching field, inspiring subsequent series of acceleration techniques based on numerical methods.

As a follower of \textit{TaylorSeer}, \textit{AB-Cache}~\cite{yu2025ab} provides mathematical explanations for the widely observed ``U-shaped similarity pattern'' in diffusion models from the perspective of numerical integration. This work models the denoising process of diffusion models as a numerical solution problem for ordinary differential equations, discovering linear relationships between outputs of adjacent steps: 

\begin{equation}\hat{\epsilon}_\theta(x_{t_\lambda(\lambda_o)}, t_\lambda(\lambda_o)) \approx r \cdot \hat{\epsilon}_\theta(x_{t_\lambda(\lambda_s)}, t_\lambda(\lambda_s))\end{equation}

where the scaling factor $r$ has the specific form:
\begin{equation}
r = \frac{\alpha_t h}{3 e^{-h}\alpha_t h - 2\sigma_{\lambda_t} e^{\lambda_o(e^h-1)}}
\end{equation}
Furthermore, \textit{AB-Cache} extends this relationship to $k$-th order Adams-Bashforth methods, establishing more general linear recursive relationships:
\begin{equation}
\hat{\epsilon}_\theta(x_{t_\lambda(\lambda_n)}, t_\lambda(\lambda_n)) \approx \sum_{i=1}^k c_i \hat{\epsilon}_\theta(x_{t_\lambda(\lambda_{n-i})}, t_\lambda(\lambda_{n-i}))
\end{equation}
The method approximates the integral equation of the denoising process 

\begin{equation}I = \int_{\lambda_s}^{\lambda_t} e^{-\tau}\hat{\epsilon}_\theta(x_{t_\lambda(\tau)}, t_\lambda(\tau)) d\tau\end{equation}

 achieving the transformation from heuristic caching to numerical integration methods and mathematically formalizing existing observational phenomena.

\textit{HiCache}~\cite{feng2025hicache} tackles the numerical instability of high-order predictive caching by identifying and overcoming fundamental limitations of traditional Taylor expansion in diffusion feature prediction.
It reveals that although Taylor polynomials provide accurate local approximations, their power-function basis leads to \textit{instability} when modeling oscillatory features, causing extrapolation overshoot and error amplification in long-timestep or high-order predictions.
Moreover, Taylor expansion lacks awareness of local geometric structures, limiting its ability to capture complex nonlinear dynamics in diffusion processes.

To address these issues, \textit{HiCache} adopts Hermite polynomials as the prediction basis, leveraging their mathematical rigor and physical relevance.
As orthogonal functions describing harmonic oscillators in quantum mechanics, Hermite polynomials inherently handle oscillatory signals effectively.

Furthermore, \textit{HiCache} introduces a contraction factor $\sigma$ to control polynomial decay, softly suppressing high-order terms while preserving expressiveness and mitigating numerical oscillations.
The prediction is formulated as: 

\begin{equation}\hat{F}_{t-k} = F_t + \sum_{i=1}^{N} \frac{\Delta^i F_t}{i!} \cdot \tilde{H}_i(-k)\end{equation}

where $\tilde{H}_i(x) = \sigma^i H_i(\sigma x)$ represents Hermite polynomials with contraction factor. 

While previous predictive caching methods primarily focus on short-term feature extrapolation, they often suffer from accumulated errors and instability under large step sizes. \textit{FoCa}~\cite{zheng2025forecast} first explicitly treats feature caching as a numerical integration problem for feature ODEs. Traditional methods (direct reuse or single-step Taylor extrapolation) only use local difference information from the most recent step, prone to extrapolation overshoot, noise amplification, and quality collapse during long jumps and ``\textit{stiff}'' phases. 

\textit{FoCa}'s technical innovation is embodied in its carefully designed two-stage process. First, in \textit{the prediction phase}, it employs the second-order Backward Difference Formula (BDF2) for multi-step extrapolation:
\begin{equation}
\hat{F}_{k+1}=\frac{4}{3}F_k-\frac{1}{3}F_{k-1}+\frac{2h}{3}\widehat{F}'_k
\end{equation}
This design fully utilizes two-step historical information, providing stronger noise resistance and numerical stability compared to single-step methods. Subsequently, in \textit{the correction phase}, it employs the Heun method for trapezoidal integration correction: $F_{k+1}=F_k+\frac{h}{2}\big(\widehat{F}'_k+\widehat{F}'_{k+1}\big)$
Although this correction step has minimal computational overhead, it significantly suppresses extrapolation overshoot and cumulative errors.

\textit{FreqCa} introduces a frequency-domain perspective to address the inherent trade-off between feature similarity and temporal continuity in diffusion model caching. 
Through detailed analysis, it reveals that the low- and high-frequency components of diffusion features exhibit distinct temporal dynamics: 
low-frequency components maintain high similarity across adjacent timesteps, whereas high-frequency components demonstrate strong temporal smoothness. 
Motivated by this observation, \textit{FreqCa} performs frequency decomposition  using a generic frequency transform $\mathcal{D}(\cdot)$:
\begin{equation}
\mathbf{z}_t = \mathbf{z}_t^{\text{low}} + \mathbf{z}_t^{\text{high}}, \quad \text{where} \quad \mathbf{z}_t^{\text{low/high}} = \mathcal{P}_{\text{low/high}}\!\big(\mathcal{D}(\mathbf{z}_t)\big).
\end{equation}
The low-frequency components $z_t^{\text{low}}$ exhibit high cross-step similarity and are thus directly reused:
\begin{equation}
\hat{z}_t^{\text{low}} = z_{t-k}^{\text{low}}, \quad k \in \{1,2,\dots,K\}.
\end{equation}
In contrast, the high-frequency components $z_t^{\text{high}}$ display smooth temporal evolution and are predicted using a second-order Hermite polynomial:
\begin{equation}
\hat{h}_i(s_t) = h_i(s_{t-1}) + \frac{\Delta s}{2}\left[h_i'(s_{t-1}) + h_i'(s_t)\right].
\end{equation}
This frequency-decoupled design bridges the advantages of the ``Cache-Then-Reuse'' and ``Cache-Then-Forecast'' paradigms, establishing a unified caching framework that achieves efficient computation without compromising generation fidelity.

Furthermore, \textit{FreqCa} introduces the \textbf{Cumulative Residual Feature (CRF)} mechanism, which aggregates features across residual connections to drastically reduce memory consumption. 
Formally, CRF is defined as:
\begin{equation}
\phi_L(x_t) = x_t + \sum_{l=1}^L \mathcal{F}_l(h^{(l)}),
\end{equation}
which reduces memory complexity from $\mathcal{O}(L)$ to $\mathcal{O}(1)$ and minimizes the number of frequency (inverse) transformations required during inference.

 \textit{FreqCa} achieves up to $7.14\times$ inference acceleration and $99\%$ memory savings on Qwen-Image,  while maintaining comparable generation quality. 


Beyond numerical prediction methods, \textit{DiCache}~\cite{bu2025dicache} employs shallow features as lightweight probes, using feature variation $\Delta_{\text{probe}}^t$ to trigger recomputation and linear interpolation for multi-step cache fusion. \textit{FastCache}~\cite{liu2025fastcache} adopts a dual strategy: it identifies static tokens via temporal significance scoring and approximates them using linear transformation $\mathbf{H}_t^s = \mathbf{W}_c \mathbf{X}_t^s + \mathbf{b}_c$; when relative change $\delta_{t,l}$ satisfies statistical criteria, it skips transformer blocks and replaces them with learnable linear transformations $\mathbf{H}_{t,l} = \mathbf{W}_l \mathbf{H}_{t,l-1} + \mathbf{b}_l$. These methods demonstrate that intelligent computation reuse across feature, token, and block levels can reduce computational overhead while maintaining generation quality, providing complementary approaches to numerical prediction paradigms.

\subsubsection{Hybrid Caching Methods}

Existing caching paradigms, such as timestep-adaptive and predictive caching, typically focus on a single optimization dimension—either temporal scheduling or numerical forecasting.  
However, diffusion inference involves complex interactions across timesteps, network layers, and feature dynamics, making single dimensional caching strategies insufficient to fully exploit the intrinsic redundancy of the process.

\textbf{\textit{Hybrid Caching Methods}} address these limitations by \textit{jointly modeling multiple dimensions: timestep, network hierarchy, and feature dynamics, within a unified framework.  }
By integrating complementary strategies such as adaptive scheduling, predictive estimation, and structural selection, these methods dynamically coordinate \emph{where and when to compute/reuse} across both spatial and temporal domains.  
This unified design enables flexible and robust cache reuse, significantly improving stability and generation quality under aggressive acceleration settings.

Caching mechanisms that depend solely on temporal similarity often fail under high acceleration ratio, as feature discrepancies between adjacent timesteps expand rapidly with larger step sizes. To address this issue, \textit{ClusCa}~\cite{zheng2025compute} introduces spatial token similarity as an orthogonal complement to temporal similarity. After completing a full computation, the model performs K-Means clustering on the token representations of the final layer, obtaining \(K\) clusters where each cluster represents a spatial region. During cache reuse, only a subset of representative tokens within each cluster are recomputed, and the results are fused with the cached features from the previous timestep to approximate updates for other tokens within the same cluster:
\begin{equation}
C(x_i)=
\begin{cases}
\mathcal{F}(x_i), & i \in I_{\text{Compute}} \\
\gamma \cdot \mu_{(i)} + (1-\gamma) C(x_i), & i \notin I_{\text{Compute}}
\end{cases}
\end{equation}
where $\mu_{(i)}$ is the mean value of computed feature of tokens in $I_{\text{Compute}}$,  formally expressed as:
\begin{equation}
    \mu_{(i)}=\frac{\sum_{j \in I_{\text{Compute}}} \mathcal{F}(x_j)}{\sum_{j \in I_{\text{Compute}}}\,[\mathbf{I}_j=\mathbf{I}_i]}.
\end{equation}

This spatiotemporal fusion mechanism significantly improves cache utilization, reducing computation while  even enhancing generation quality.

Beyond spatial aggregation, another class of hybrid strategies approaches caching from the \emph{Forecast-Then-Verify} perspective, aiming to balance extrapolation efficiency and error control in multi-step inference. Inspired by speculative decoding in large language models, \textit{SpeCa}~\cite{liu2025speca} constructs a dual-phase closed-loop framework of prediction and verification. Specifically, a lightweight draft predictor (e.g., TaylorSeer) extrapolates the next \(k\) steps via a Taylor-series expansion:
\begin{equation}
\mathcal{F}_{\text{pred}}(x_{t-k}^l)=\mathcal{F}(x_t^l)+\sum_{i=1}^{m}\frac{\Delta^i \mathcal{F}(x_t^l)}{i!\,N^i}(-k)^i,
\end{equation}
where \(\Delta^i\) denotes the \(i\)-th order finite difference. The model then computes the relative error between the predicted and true values through a lightweight verifier:
\begin{equation}
e_k=\frac{\lVert \mathcal{F}_{\text{pred}}(x_{t-k}^l)-\mathcal{F}(x_{t-k}^l)\rVert}{\lVert \mathcal{F}(x_{t-k}^l)\rVert},
\end{equation}
and compares it with a dynamic threshold \(\tau_t\). If \(e_k \le \tau_t\), the prediction is accepted; otherwise, the model rolls back to the last verified state and performs a full computation for correction. This design acts as a ``firewall'', enabling aggressive step extrapolation while maintaining stability. The theoretical acceleration ratio can be approximated as
\begin{equation}
\mathcal{S}\approx \frac{1}{(1-\alpha)+\gamma},
\end{equation}
where \(\alpha\) is the prediction acceptance rate and \(\gamma\) is the verification cost ratio (typically much smaller than $1$). \textit{SpeCa} effectively suppresses error accumulation under high acceleration, offering a reliable extrapolation mechanism for inference.

In contrast to heuristic methods relying on local feature similarity, \textit{OmniCache}~\cite{chu2025omnicache} revisits cache scheduling from a global sampling-trajectory perspective. Empirical observations reveal that feature trajectories of diffusion models exhibit a ``\textit{boomerang}'' shape in low-dimensional space, implying that model states evolve more smoothly in certain low-curvature stages, making them suitable for cache reuse. \textit{OmniCache} thus introduces a curvature-guided global scheduling strategy, reusing caches during low-curvature phases to minimize interference, while employing a noise correction mechanism:
\begin{equation}
\begin{aligned}
q_{\theta}(x_t,t) & =\tilde{\epsilon}_{\theta}(x_t,t)-\epsilon_{\theta}(x_t,t) \\
q_{\theta}(x_{t-1},&t-1)\approx \gamma_{t-1}\,q_{\theta}(x_t,t),
\end{aligned}
\end{equation}
where \(q_{\theta}\) denotes cache-induced noise. The model selectively applies low-pass or high-pass filtering to this noise depending on the sampling stage—preserving global structure in early phases and enhancing fine-grained detail later. Notably, this method requires no additional training, combining offline calibration with online reuse.

In the task- and structure-adaptive direction, researchers have further integrated caching with importance estimation, regional sparsity computation, and multi-frame compression.  
\textit{TokenCache}~\cite{lou2024token} targets the DiT architecture by using a lightweight predictor to evaluate token importance and aggregating tokens into blocks, where only low-importance blocks undergo cache reuse, thereby improving inference efficiency without sacrificing detail.  
\textit{FISEdit}~\cite{yu2024accelerating} focuses on image editing tasks, leveraging an automatic masking mechanism (based on latent differences and OTSU thresholding) to locate affected regions. It employs pixel-level sparse convolution and attention to recompute only modified areas, while using asynchronous pipelined caching to reduce memory usage.  
For video generation, \textit{FlexCache}~\cite{sun2024flexcache} identifies strong cross-frame feature correlations, applying keyframe selection and linear interpolation for two-stage compression. It further employs object–background decoupled retrieval and an LRBU (Least Recently Beneficial Used) replacement policy to improve cache hit rate and timeliness. Together, these methods embody the hybrid principle of ``\textit{multi-dimensional coordination + dynamic adaptivity.}''

From a structural standpoint, \textit{RainFusion}~\cite{chen2025rainfusion} optimizes video diffusion acceleration through sparse attention. It identifies that conventional binary partitioning of attention heads (spatial vs.\ temporal) overlooks a third category: \emph{textural heads}, which are crucial for modeling high-frequency semantics. \textit{RainFusion} thus introduces a ternary attention classification and an Adaptive Recognition Module (ARM) to dynamically determine head types online. The ARM evaluates the recall ratio under masked attention:
\begin{equation}
R'=\frac{S(Q',K',M')}{S(Q',K',M_{\mathrm{init}})},
\end{equation}
and selects the corresponding sparsity pattern when \(R'\) exceeds a threshold; otherwise, it employs checkerboard sampling to preserve detail. \textit{RainFusion} achieves low-cost adaptive coupling of sparsity and caching, significantly improving efficiency without degrading fidelity.

From another perspective, \textit{ProfilingDiT}~\cite{ma2025model} approaches caching from a semantic perspective, revealing long-term attention biases between foreground and background across DiT layers. Based on this, it proposes a dual-granularity adaptive caching framework.  
Across layers, it classifies modules into a foreground set \(F\) and background set \(B\) according to the foreground attention ratio:
\begin{equation}
R_{\mathrm{attn}}=\frac{N_{\mathrm{high}\cap fg}}{N_{fg}}.
\end{equation}
Across timesteps, it dynamically adjusts the cache interval:
\begin{equation}
T_s=T_{\max}-(T_{\max}-T_{\min})\cdot\frac{s-s_0}{S-s_0},
\end{equation}
realizing a progressive strategy of coarse caching in early stages and fine updates later.

\textit{BlockDance}~\cite{zhang2025blockdance} analyzes the structural heterogeneity of DiT and observes that shallow and middle-layer features change minimally during late denoising stages, while deep-layer features continue to evolve significantly. This reflects the ``structure-first, detail-later'' principle of diffusion generation: early stages construct global layouts, and later ones refine textures. Accordingly, \textit{BlockDance} proposes a hierarchical caching strategy that maintains full computation during early stages for quality assurance, while reusing shallow features at fixed frequencies in later stages to eliminate redundancy.  
Further, \textit{BlockDance-Ada} introduces a lightweight decision network that leverages latent variable \(z_p\) and text embedding \(c\) to dynamically assess content complexity and generate an instance-specific reuse strategy \(u\). Its objective is defined as:
\begin{equation}
R(u)=C(u)+\lambda Q(u),\quad C(u)=1-\frac{\sum_{t=1}^{S-\rho}u_t}{S-\rho}.
\end{equation}
balancing acceleration gain and generation quality, thereby enabling instance-level adaptive caching.

\textit{HyCa}~\cite{zhengLetFeaturesDecide2025} introduces a novel perspective on predictive caching by addressing the key limitation of prior approaches that assume uniform temporal dynamics across all hidden dimensions of DiTs. In practice, diffusion features exhibit highly heterogeneous temporal behaviors---some dimensions undergo oscillatory or multimodal evolution, while others vary smoothly and predictably over time. To capture this intrinsic heterogeneity, \textit{HyCa} models feature evolution as a \emph{mixture of ordinary differential equations (ODEs)}, enabling distinct subsets of feature dimensions to follow independent dynamic trajectories.

Concretely, \textit{HyCa} performs unsupervised clustering of feature dimensions based on dynamic descriptors such as acceleration ratios and curvature ratios, grouping dimensions with similar temporal behaviors. Given a solver set \( \mathcal{S} \), each cluster \( c \) is assigned its optimal solver \( s_c^\star \) by minimizing the mean next-step prediction error within the cluster:
\begin{equation}
\min_{\{s_c \in \mathcal{S}\}_{c=1}^C}
\sum_{c=1}^{C}
\left[
  \frac{1}{|c|}
  \sum_{d \in c}
  \left\|
  \hat{\mathcal{F}}_{t+1}^{(s_c,d)} - \mathcal{F}_{t+1}^{(d)}
  \right\|_2^2
\right],
\end{equation}
where \( \hat{\mathcal{F}}_{t+1}^{(s_c,d)} \) denotes the predicted feature value of dimension \( d \) at timestep \( t+1 \) using solver \( s_c \). Remarkably, these clustering assignments remain stable across prompts, resolutions, and timesteps, enabling a ``\textit{One-Time Choosing, All-Time Solving}'' inference paradigm in which solver selection is performed once offline and reused throughout inference without additional overhead. By combining complementary solvers, \textit{HyCa} balances stability and adaptability, aligning caching dynamics with heterogeneous feature evolution. This mixture-of-ODE framework offers a unified, training-free approach that improves both efficiency and fidelity in diffusion inference.

Overall, hybrid caching methods achieve fine-grained and intelligent cache reuse by integrating cooperative mechanisms across temporal, spatial, structural, and semantic dimensions. Rather than relying on simple timestep skipping, they dynamically adjust reuse patterns guided by feature evolution, structural divergence, and content semantics. Consequently, these strategies balance stability and visual fidelity under high acceleration, providing new theoretical and practical pathways for accelerating diffusion model inference.

\section{Applications of Cache Acceleration}

With the rapid development of diffusion models in multimodal generation tasks, cache mechanisms have emerged as an essential strategy for improving inference efficiency. The core idea is to reuse intermediate features across time steps or spatial regions to avoid redundant computation, thereby achieving significant acceleration without sacrificing generation quality. Cache strategies have demonstrated strong adaptability and generality across various downstream applications, including image and video editing, 3D generation, speech synthesis, super-resolution reconstruction, world model construction, discrete diffusion models and AI for Science category. Depending on the task, cache techniques often exploit \textbf{spatial locality}, \textbf{temporal redundancy}, or \textbf{feature consistency} to realize task-specific computation reuse and acceleration.

\subsection{Acceleration in Image and Video Editing Tasks}
In controllable image and video generation tasks,  a large amount of input such as background areas usually remains unchanged, but consumes a lot of computation. By leveraging spatial and temporal redundancy, these areas can be efficiently cached and reused. For image editing, caching intermediate features of unedited regions and reusing them in subsequent iterations is a natural acceleration strategy. \textit{EEdit}~\cite{eedit} adopts this idea by caching and selectively updating intermediate results of self-attention and MLP layers based on spatial locality (SLoC), thereby avoiding redundant computation without compromising the fidelity of edited regions.  
When control signals (e.g., from ControlNet~\cite{zhang2023adding}) are introduced, similar redundancies exist across both spatial and temporal dimensions. For instance, \textit{EVCtrl}~\cite{evctrl} divides the network into global and local functional regions, caching the global uncontrolled features and recomputing only the locally controlled ones while performing sparse temporal updates at key frames or denoising steps. This approach achieves more than a $2\times$ speed-up in tasks like CogVideo-ControlNet with negligible quality loss.  
In controllable video generation, adjacent frames also exhibit high temporal similarity. \textit{Follow-Your-Emoji-Faster}~\cite{fyef} caches features at key time steps and reconstructs intermediate frame features via Taylor interpolation, avoiding full inference for each frame. Furthermore, it employs facial landmarks to guide update frequency in spatial regions, minimizing cumulative errors. This face-guided interpolation caching achieves around $2.6\times$ lossless acceleration while preserving fine details. Collectively, these studies highlight the plug-and-play nature and strong generality of cache mechanisms in editing tasks.

\subsection{Acceleration for 3D Generation Tasks}
In 3D generation, both neighboring camera views and adjacent time steps contain strong redundancy, enabling cache-based acceleration without retraining. \textit{Hash3D}~\cite{hash3d} employs a hash-bucket-based feature reuse strategy that shares features between nearby viewpoints, reducing inconsistencies from independent noise sampling and improving smoothness and coherence of generated 3D models. This approach achieves up to a $4\times$ speed-up in DreamGaussian (from 2 minutes to 30 seconds) and a $3\times$ acceleration in Zero-123 (from 20 minutes to 7 minutes). In text-to-3D generation, the speed-up ranges from $1.5\times$ to $1.9\times$, while PSNR and SSIM show slight improvements and LPIPS exhibits minor degradation. Importantly, CLIP-G scores remain comparable to the baseline, indicating preserved semantic consistency.

\subsection{Acceleration in Audio Generation Tasks}
Speech generation tasks can also benefit from training-free cache acceleration. \textit{DiTReducio}~\cite{DiTReducio} identifies two types of redundancy: temporal redundancy where model outputs across adjacent denoising steps are highly similar, and branch redundancy where conditional and unconditional branches in CFG exhibit near-identical outputs. By computing only one branch and reconstructing the other using cached residuals, DiTReducio achieves more accurate results than direct reuse. For TTS caching, a progressive calibration process is employed to determine whether the reuse-induced loss is below a predefined threshold, thereby ensuring high cache precision. Experiments show that this approach reduces computation by $50\%$ and improves inference speed by $30\%$.

\subsection{Acceleration in Super-Resolution Tasks}
Image super-resolution, as one of the canonical diffusion model applications, also exhibits temporal redundancy in its denoising process, allowing for training-free cache-based acceleration. \textit{HiCache}~\cite{feng2025hicache} advances this direction by introducing a Hermite-polynomial-based interpolation caching scheme on the InfDiT model. The method achieves a theoretical acceleration of up to $5.93\times$ and a practical wall-clock speed-up of $2.43\times$, while outperforming TaylorSeer~\cite{liu2025reusing} in SSIM and PSNR on the NTIRE 2025~\cite{ntire2025srx4} benchmark.

\subsection{Acceleration in World Models}
Recently, video-based world models have become a cornerstone of generative video modeling and interactive content generation. These models often follow a ``\textit{base diffusion model training + control signal integration → autoregressive distillation}'' paradigm. High-fidelity diffusion backbones (e.g., Wan 2.1/2.2~\cite{wan2025}) are trained on diverse video data and integrated with external control signals such as mouse, keyboard, or environment parameters. Subsequently, the model is distilled into a few-step autoregressive generator for efficient inference. During this process, caching modules~\cite{liu2025timestep, song2025hero} are integrated into diffusion backbones to cache cross-timestep features and reduce computational costs. Furthermore, when the model is distilled to only 4–10 inference steps, accuracy degradation and temporal instability can be alleviated via \textit{Diffusion Caching}, which reuses intermediate features from adjacent time steps. This enables significant computational savings without modifying model parameters or structure, ensuring both efficiency and temporal consistency in world model generation, as demonstrated in systems like Google Genie 3~\cite{genie3world}, GameCraft~\cite{li2025hunyuangamecrafthighdynamicinteractivegame}, and Matrix-Game 2.0~\cite{MatrixGame2.0-2025}.

\subsection{Acceleration in Discrete Diffusion Models}
Interestingly, even for diffusion-based discrete diffusion models, caching can effectively exploit hidden redundancies to achieve acceleration. This includes unified multimodal understanding and generation, and discrete diffusion language models (dLLMs).  
\textit{Lumina-DIMOO}~\cite{xin2025luminadimooomnidiffusionlarge} achieves state-of-the-art performance in unified multimodal understanding, generation, and editing tasks, while improving sampling efficiency via a Max-Logit-based Cache (ML-Cache) mechanism, doubling sampling efficiency. Similarly, in the rapidly evolving diffusion language modeling domain, \textit{dLLM-Cache}~\cite{liu2025dllm} accelerates inference by combining long-interval prompt caching, short-interval response caching, adaptive partial updates, and a V-verify mechanism. On the LLaDA-8B~\cite{nie2025large} Instruct model under the GPQA~\cite{rein2024gpqa} benchmark, this achieves an $8.08\times$ speed-up, reducing FLOPs per token from 22.07T to 2.73T. As the diffusion paradigm continues to converge with diverse fields, cache-based acceleration techniques are expected to gain increasing attention and broader adoption.

\subsection{Acceleration in AI for Science}
In molecular geometry generation, flow-matching models typically rely on ODE solvers requiring multiple iterative steps. Since molecular coordinates and atom types evolve smoothly across iterations, intermediate neural features exhibit temporal redundancy that can be exploited via predictive caching. The \textit{AB-Cache}~\cite{Sommer2025PredictiveFC} achieves with more than $3\times$ acceleration while getting competitive results compared to 100-step full baselines across multiple metrics, including energy, strain, validity, and molecular stability on GEOM-Drugs~\cite{axelrod2022geom} dataset. This demonstrates that cache-based acceleration holds substantial promise for AI4Science domains, offering efficient and accurate computation for scientific generative modeling tasks.

\section{Future Perspectives of diffusion Caching}


In recent years, cache-based acceleration techniques have emerged as a promising solution for speeding up diffusion model inference by reusing intermediate representations across successive steps. However, this approach inherently incurs a trade-off between \textbf{computational redundancy} and \textbf{memory overhead}. Specifically, three core challenges remain to be addressed: first, the substantial memory footprint required to store intermediate activations, second, potential degradation in generative quality caused by cache-induced approximation errors and third, a lack of theoretical grounding and systematic integration with complementary acceleration strategies. In the following, we discuss each of these challenges in turn and outline potential directions for future research.

\subsection{Memory Consumption Challenge}
\label{sec:5.1}


Cache-based mechanisms can substantially accelerate the diffusion process by reusing intermediate representations and thereby eliminating redundant computations. However, this efficiency gain comes at the cost of increased memory consumption. In particular, activations that would normally be released after each step must be retained across successive diffusion steps for either direct reuse or partial recomputation, leading to a significant GPU memory overhead. While such caching strategies effectively enhance computational efficiency, they also place a substantial burden on the memory subsystem.

Large diffusion models today often contain billions of parameters, requiring over 10~GB of GPU memory solely for parameter storage. For instance, Hunyuan-Image 3.0~\cite{cao2025hunyuanimage} comprises an enormous 80B parameters, which cannot even fit within a single 80~GB GPU. When cache storage is taken into account, the overall memory footprint increases substantially. In particular, cache memory consumption typically \textbf{scales linearly} with the token sequence length. In a $512\times512$ image generation task, the KV-cache alone can demand an additional 4–8~GB of memory. Under long-sequence or memory-constrained settings, this often results in \textbf{Out-of-Memory (OOM)} errors. Moreover, state-of-the-art methods such as \textit{TaylorSeer} further exacerbate this issue by caching activations for every network layer, causing memory usage to grow linearly with network depth and introducing an additional 7–8~GB of GPU overhead.

Such substantial memory demands present serious challenges for hardware deployment. On mobile devices (e.g., smartphones or tablets) with only 4-12~GB of unified memory, it becomes infeasible to load both model parameters and cache data concurrently. In embodied AI and robotics applications—where power consumption and chip area are tightly constrained—the high memory requirements of caching frequently exceed available hardware capacity. Even in high-resolution image generation or video synthesis tasks on high-end GPUs (e.g., NVIDIA A100 80~GB), cache usage scales approximately quadratically with sequence length, creating critical memory bottlenecks. Furthermore, in multi-task concurrent inference scenarios, memory contention among parallel jobs exacerbates this problem and degrades overall throughput.

To alleviate these issues, researchers have proposed several memory-efficient caching strategies. A representative example is \textit{FreqCa}~\cite{liu2025freqca}, which introduces a \emph{Memory-Efficient Feature Caching} scheme. Traditional cache mechanisms store all features from both the attention and feed-forward (FFN) layers, leading to exorbitant memory costs (e.g., \textit{ToCa} requires more than 10~GB on FLUX). \textit{FreqCa}, inspired by the residual-network interpretation that residual connections capture an ensemble of features across layers, proposes the \textbf{Cumulative Residual Feature (CRF)} caching strategy. By retaining only the cumulative residual feature vector, this method compresses the $2\times L$ layer features into a single representation, reducing cache memory usage by up to 99\%. Moreover, it cuts the number of frequency-decomposition operations by roughly $2L$, such that the total latency introduced by caching remains below 0.01\% of the overall diffusion process. As a result, \textit{FreqCa} markedly enhances the memory efficiency and practical deployability of cache-based acceleration techniques.

\subsection{Generation Quality Degradation}
\label{sec:5.2}

Cache-based acceleration improves inference efficiency by reusing historical intermediate results, thereby reducing redundant computation. However, this reuse inevitably introduces \textbf{cache-induced errors}, which can lead to the loss of fine-grained information during the diffusion process. As some intermediate computations are skipped or approximated, the model’s ability to capture complex feature dependencies is weakened, resulting in degraded generation quality. Common artifacts include texture blurring, edge distortion, and loss of micro-structures. At higher acceleration ratios, these errors accumulate progressively, leading to noticeable deterioration in visual fidelity.

Such quality degradation is particularly problematic in detail-sensitive or high-precision tasks. For example, in facial generation or recognition, minor inaccuracies in local regions (e.g., eye corners or nose wings) can cause feature mismatches; in medical imaging or scientific visualization, even slight distortions in tumor boundaries or vessel structures can significantly impact diagnostic accuracy. Consequently, although cache-based diffusion acceleration is highly practical for amusement-level AIGC tasks such as visual effects, artistic creation, and digital humans, its applicability to high-precision generation remains limited. Current cache mechanisms are better suited for applications with moderate detail constraints, while fields such as medical imaging, security recognition, and engineering design still require solutions that balance acceleration with generation quality.

\subsection{Theoretical Limitations and Future Directions}
\label{sec:5.3}

\subsubsection{Lack of Theoretical Frameworks}
\label{sec:5.3.1}

Although cache methods have achieved remarkable success in accelerating diffusion model inference, with some SOTA approaches such as \textit{TaylorSeer} reaching nearly $5\times$ acceleration. Their development remains largely empirical. Existing techniques are primarily \textbf{engineering-driven} explorations rather than theories grounded in mathematical rigor. The distributional deviation introduced during cached inference lacks interpretable and verifiable theoretical characterization. Current quality assurance largely relies on empirical parameter tuning rather than controllable, theoretically guided analysis, which is inefficient and lacks generalization across different tasks such as image generation, video prediction, and 3D modeling.

Some studies have attempted to investigate the theoretical underpinnings of cache-induced errors. For instance, \textit{ERTA Cache}~\cite{peng2025ertacache} categorized these errors into two main types: \textit{Feature Shift} and \textit{Step Amplification}, providing preliminary insight into how caching affects the diffusion process. However, such analyses remain heuristic and lack systematic mathematical derivation or unified theoretical frameworks. Their conclusions often depend on empirical observations rather than first-principle reasoning from diffusion dynamics, thus providing limited theoretical guidance for future algorithm design.

Furthermore, little attention has been given to understanding how caching interacts with different sampling strategies such as DDIM or Flow Matching. The absence of a unified framework for analyzing how cache strategies can be adapted to diverse samplers, minimize error propagation, and enhance sampling efficiency has restricted their generalization across generation paradigms. This theoretical gap not only limits the adaptability of caching in various diffusion settings but also constrains its integration with other efficient inference techniques. Therefore, future research should develop a more rigorous, unified theory of cache-induced error analysis, quantitatively characterizing its effect on diffusion dynamics and exploring the compatibility of cache strategies with different sampling approaches. Such theoretical advancements would lay a solid foundation for high-performance and reliable diffusion inference.

\subsubsection{Integration with Other Acceleration Strategies}
\label{sec:5.3.2}

Cache strategies possess strong flexibility and orthogonality, making them highly compatible with other model acceleration techniques. When integrated with mainstream methods such as \textbf{model distillation}, \textbf{pruning}, and \textbf{quantization}, caching can further enhance inference speed while maintaining generation quality and reducing memory footprint, achieving a balanced trade-off between performance and efficiency.

Recent works have explored the synergy between caching and other acceleration techniques. For example, \textit{DaTo}~\cite{zhang2024token} combines caching with token pruning to jointly improve inference speed and memory efficiency through dynamic token selection and feature reuse. \textit{QuantCache}~\cite{wu2025quantcache} systematically studies the fusion of caching and quantization, achieving significant memory savings without sacrificing output quality. \textit{Jenga}~\cite{zhang2025training} introduces an integrated framework that combines \textit{TeaCache}~\cite{xin2025luminadimooomnidiffusionlarge} with variable-resolution inference, enabling multi-scale acceleration with improved efficiency. These works collectively demonstrate that cache mechanisms not only extend their own acceleration potential but also complement other optimization paradigms, offering promising solutions for achieving high-quality generation under limited memory budgets.

Nevertheless, achieving effective coordination among multiple acceleration techniques remains a major challenge. Most existing methods still face a trade-off between inference speed and generation quality: aggressively pursuing speed may amplify approximation errors and degrade output fidelity, whereas prioritizing quality preservation often reduces computational efficiency. When multiple acceleration mechanisms are applied jointly, their induced errors may compound, thereby exacerbating distributional deviations from the original model. Therefore, it is essential to design principled integration frameworks that can mitigate or counterbalance these accumulated errors rather than allowing them to linearly aggregate. The core issue lies in the insufficient theoretical understanding of error propagation and interaction across different acceleration mechanisms. Future research should aim to establish \textbf{a unified analytical framework} that quantitatively models these interactions and identifies the optimal conditions for deep integration among caching, quantization, pruning, and other acceleration strategies. Such efforts will not only guide the design of robust and efficient diffusion inference systems but also foster theoretical advancements that bridge \textbf{engineering practice} with \textbf{diffusion theory}.

\section{Conclusion}
In summary, Diffusion Models (DMs) have achieved remarkable progress in both generation quality and controllability, becoming one of the key foundations of modern generation models. However, the high computational cost and long inference time remain key challenges that hinder their widespread adoption in real-world applications. In scenarios such as real-time interaction, on-device deployment, or large-scale online generation, the heavy inference load significantly limits the model’s practicality, scalability, and cost-efficiency. Existing acceleration methods, including model distillation, pruning, sampler optimization, and system-level operator acceleration, have shown some success in specific cases. However, these methods often suffer from limited applicability, degraded generation quality, or high training costs, making it difficult to achieve efficient and stable acceleration in a generalizable manner. 
In contrast, \textbf{\textit{Diffusion Caching}} offers a new training-free and model-agnostic approach to speeding up inference. Its main idea is to identify and reuse redundant computations in the diffusion process, thus cutting computational costs without harming output quality. This enables a lightweight, efficient, and composable form of inference optimization. Especially with the rise of the ``\textit{Cache-Then-Forecast}'' approach, which represented by TaylorSeer, diffusion caching has achieved lossless acceleration on mainstream models with high speedup ratios, showing great promise and broad applicability.
Looking ahead, as model sizes grow and generative tasks become more multimodal and interactive, diffusion caching may become a key part of next-generation efficient generative frameworks. It could evolve into a standard module in diffusion inference pipelines and drive joint optimization across algorithmic and system levels. This will help support efficient generation for images, videos, 3D content, and even world models.
From a broader view, the idea of  \textbf{\textit{Diffusion Caching}} may become an important path toward achieving truly scalable and sustainable generative intelligence.

\bibliographystyle{IEEEtran}
\bibliography{references}

\end{document}